\PassOptionsToPackage{table}{xcolor}
\PassOptionsToPackage{most}{tcolorbox}
\documentclass[runningheads]{llncs}

\usepackage{eccv}

\usepackage{eccvabbrv}
\usepackage{tabularray}

\usepackage{dblfloatfix}
\usepackage{graphicx}
\usepackage{booktabs}
\usepackage{wasysym}
\usepackage{gensymb}
\usepackage{amsmath}
\usepackage{amssymb}
\usepackage{multirow}
\usepackage{makecell}
\usepackage{tcolorbox}

\usepackage[colorlinks=true, 
     linkcolor=blue, 
     filecolor=blue, 
     citecolor =blue,       
     urlcolor=blue]{hyperref}

\usepackage{orcidlink}
\providecommand{\authcount}[1]{} 

\begin{document}

\title{Residual-Guided Expert Specialization \\for Incomplete Multimodal Learning} 

\titlerunning{MARS: Missingness-Aware Residual-guided Specialization}

\author{Seunghun Baek\orcidlink{0009-0008-9514-407X} \and
Jihwan Park\orcidlink{0009-0004-0179-2166} \and
Jaeyoon Sim\orcidlink{0009-0001-8696-3849} \and
Minjae Jeong\orcidlink{0009-0003-3442-889X} \and
\\Hoseok Lee\orcidlink{0009-0004-1306-676X} \and
Won Hwa Kim\orcidlink{0000-0001-5393-0883}}

\authorrunning{S. Baek et al.}

\institute{Pohang University of Science and Technology (POSTECH), South Korea\\
\textbf{Code:} \url{https://github.com/seunghub/MARS}}

\makeatletter
\let\mars@origaddcontentsline\addcontentsline
\let\addcontentsline\@gobblethree
\makeatother
\maketitle
\vspace{-2mm}
\begin{abstract}
\sloppy
As real-world prediction systems often face missing modalities at inference, 
incomplete multimodal learning (IML) remains a practical challenge. 
While prior methods aim to learn representations robust to missing inputs, 
representations from incomplete modalities inevitably deviate from their full-modality counterparts due to missing evidence. 
To explicitly leverage these 
deviations,
we propose \textbf{MARS} (\textbf{M}issingness-\textbf{A}ware \textbf{R}esidual-guided \textbf{S}pecialization), a mixture-of-experts framework that 
guides expert specialization based on 
how representations are reshaped by missingness.
By contrasting task representations derived from incomplete inputs with their complete counterparts during training,
we derive a privileged residual signal that captures 
this representational gap.
The residual signal guides \textbf{a residual router} to assign samples to the experts specialized for the corresponding deviation patterns.
In parallel, \textbf{a feature router} learns to imitate this routing behavior using only incomplete inputs, 
enabling deployment without access to full modalities.
To mitigate this train–test router gap, we develop a discrepancy-aware noise regularization that adaptively perturbs the residual router’s decisions when the feature router deviates, 
enhancing the expert robustness under imperfect imitation.
Experiments on multimodal classification (CASIA-SURF, CREMA-D, UPMC Food-101) and segmentation (MCubeS) 
under missing scenarios
show that MARS consistently surpasses baselines, 
while remaining efficient and extensible to diverse backbones and tasks.
\vspace{-2mm}
\vspace{-2mm}
\end{abstract}

\vspace{-1mm}
\section{Introduction}
\label{sec:intro}


Leveraging information from multiple sources (\ie, modalities) has become a mainstream topic
in computer vision across diverse tasks (\eg, classification~\cite{multimodal_classification,CASIASURF,multimodal_classification2}, segmentation~\cite{multimodal_segmentation,MCubeS}) and domains (\eg, medical imaging~\cite{medical4,medical5,medical3}, remote sensing~\cite{remotesensing1,remotesensing2}).
By integrating complementary cues from diverse sources, 
multimodal learning enables a deeper understanding of the target task.
However, utilizing all modalities, which were available during training, may be infeasible during inference due to 
practical reasons such as sensor malfunction or acquisition cost.
To make multimodal systems practically reliable, it is crucial to make prediction models robust to test-time missingness.
This {\em incomplete multimodal learning} (IML) 
setting is distinct from scenarios~\cite{FlexMoE, train_missing} that assume missingness during training, 
whose objective is to fully leverage incomplete multimodal data.



\begin{figure}[t]
  \centering
   \includegraphics[width=0.99\linewidth]{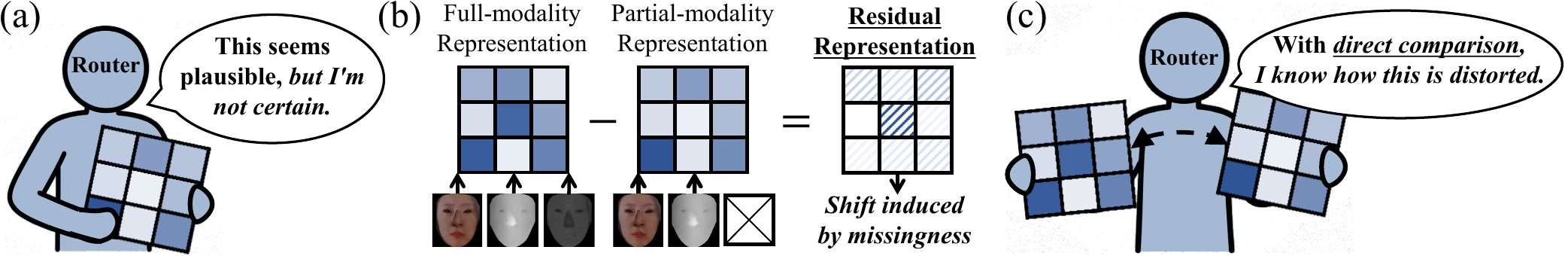}
    \vspace{-2mm}
   \caption{\small Motivation of our residual-guided routing strategy.
    (a) In conventional Mixture-of-Experts (MoE) frameworks, the router bases its decision on the same input representation as the experts, which may lack sufficient task evidence and 
    be distorted under missing modalities.
    (b) By contrasting the partial-modality representation with its full-modality counterpart, the residual captures how missingness reshapes the task representation.
    (c) Our router conditions expert specialization on this residual, enabling routing that accounts for such deviations and yields more reliable predictions.
    }
   \label{fig:concept}
   \vspace{-4mm}
\end{figure}

Under the IML scenario, 
various approaches have been proposed to handle potential modality missingness.
Imputation-based methods~\cite{imputation1,imputation3} aim to reconstruct the missing modalities from the observed ones. 
However, these approaches often suffer from data hallucination~\cite{hallucination,hallucination2} and heavy computational overhead~\cite{survey}, which limit their practicality. 
Recent efforts have therefore shifted toward learning representations that are inherently robust to potential 
modality missingness.
Early approaches, such as LCR~\cite{LCR} and RFNet~\cite{RFNet}, focus on learning modality-invariant features, 
but they unavoidably discard useful complementary cues from individual modalities. 
Subsequent studies have attempted to retain the complementary information via feature-level imputation~\cite{ShaSpec}, uncertainty modeling~\cite{DMRNet}, or expert-based architectures~\cite{MoMKE,SimMLM}.
But still, they rely solely on simulated input with missingness without explicitly exploring the {\em direct effect of the absence}, which causes systematic representation deviation. 

{\bf Key insight.} Unlike previous works, we examine how modality removal reshapes the task representation.
The representation computed from all modalities serves as the most task-sufficient reference, and
when some modalities are absent, 
the resulting representation is inevitably altered as supportive evidence is removed.
Although MMANet~\cite{MMANet} attempts to align representations from partial inputs with their full-modality counterparts via knowledge distillation,
the partial-modality representations still remain unreliable due to inevitable information loss.
Rather than relying solely on this altered representation, 
we consider {\em the difference (i.e., residual)} between representations obtained with and without missing modalities.
This residual directly reflects how the task representation shift due to modality removal, 
as illustrated in \cref{fig:concept}.
By leveraging such deviation patterns captured by the residual, 
the model can better account for deviations in the input representation instead of making blind decisions.

In this regime,
we introduce \textbf{MARS} (\textbf{M}issingness-\textbf{A}ware \textbf{R}esidual-guided \textbf{S}pecialization), a mixture-of-experts (MoE) framework~\cite{sparseMoE} equipped with 
{\em a residual router} that inputs the residual.
Exposing such residual during training allows the router to directly assign samples to experts specialized for the corresponding deviation pattern (\cref{sec:residual_router}). 
Notice that this residual router cannot operate without access to complete information.
Therefore, for test-time deployment with missing modalities, 
{\em a feature router} is introduced to approximate the residual router’s behavior without relying on residual signals (\cref{sec:feature_router}).
This {\em dual-router scheme} thus enables a similar routing strategy at inference. 

Moreover, to mitigate the inherently inevitable 
discrepancy between the residual and feature routers, 
we introduce a discrepancy-aware noise regularization (\cref{sec:noise}) that adaptively increases the residual router's stochasticity for experts with larger discrepancies between 
the feature and residual 
routers.
This mechanism enhances expert robustness by encouraging exploration of alternative experts 
that may be 
preferable at 
test time 
under {\em imperfect imitation}.
In addition, a discrepancy-guided sampling 
(\cref{sec:sampling}) dynamically prioritizes modality combinations where the routers disagree the most, facilitating balanced learning 
under incomplete conditions. 
Together, these components yield robust expert behaviors and consistent performance gains across 
diverse domains and tasks.

Overall, our contributions are summarized as threefold: 
\begin{itemize}
    \item \textbf{Residual perspective.} 
    We reformulate IML through a characterization of representation deviations in incomplete inputs, 
    where routing is conditioned on such deviations while experts operate on incomplete features. 
    This separation enables explicit expert specialization to missing-modality patterns 
    rather than relying solely on unreliable incomplete representation.

    \item \textbf{Dual-router design.} 
    We introduce a dual-router framework that disentangles expert specialization and deployment.
    A deployable feature router learns to replicate the privileged residual router's behavior at inference. 
    Beyond knowledge distillation, we further guide the routing noise to mitigate train–test routing discrepancy and promote robust expert specialization.

    \item \textbf{Efficiency and generality.} 
    Through extensive experiments on four datasets across diverse domains and tasks, 
    MARS consistently improves performance across modality combinations 
    while remaining computationally efficient.
\end{itemize}
\vspace{-3pt}
\section{Related Work}
\vspace{-3pt}
Existing IML methods for test-time modality missingness 
can be broadly 
categorized into imputation-based, representation-based, and 
expert-based approaches.

\noindent\textbf{Imputation-based methods. }
Early works attempt to impute missing inputs from existing modalities.
Generative models~\cite{imputation1,imputation3}
have been employed to reconstruct the absent modalities.
However, retrieving signals from partial observations often leads to quality issues~\cite{hallucination,hallucination2}, 
while the computational overhead and reliance on generative models at inference make real-time applications infeasible.

\noindent\textbf{Representation learning under incompleteness.}
A subsequent line of work aims to learn representations that are robust to simulated modality missingness.
HeMIS~\cite{HeMIS}, LCR~\cite{LCR} and RFNet~\cite{RFNet} learn modality-invariant representations,
whereas mmFormer~\cite{mmFormer} and ShaSpec~\cite{ShaSpec} 
retain modality-specific cues. 
While MMANet~\cite{MMANet} distills model knowledge from a full-modality teacher to a partial-modality student, 
DMRNet~\cite{DMRNet} enhances robustness through 
probabilistic uncertainty modeling.
However, the estimated uncertainty itself remains unreliable, 
as it is inferred solely from partial observations without explicit supervision.

\noindent\textbf{Expert-based specialization.}
MoMKE~\cite{MoMKE} first introduced modality-specific experts, trained on unimodal data, and aggregated them through a soft router during a second training stage.
However, all available inputs are fed into every expert during aggregation, 
which leads to unnatural cross-feeding (e.g., using RGB input in an IR-trained expert) and high computational overhead.
SimMLM~\cite{SimMLM} alleviated this by routing each modality only to its corresponding expert, 
thereby achieving more efficient inference.
Nevertheless, its aggregation operates at the logit level, which limits fine-grained specialization and 
increases computational cost.
Moreover, both methods require two-stage training and impose a structural constraint that ties the number of experts to the number of modalities.


In contrast, Flex-MoE~\cite{FlexMoE} defines experts at the modality-combination level, assigning a dedicated expert to each combination while allowing additional experts to be selected per sample. 
However, this design still requires at least $2^M$-1 experts for $M$ modalities. 
Moreover, defining experts strictly by modality combinations 
restricts specialization to predefined modality combinations. 
In practice, the task-specific contribution of each modality may vary across samples.
Samples within the same combination may require distinct expertise, 
while different combinations often exhibit similar representation deviation patterns.

\noindent\textbf{Our perspective.}
Our model neither {\em i)} imposes structural constraints or predefined roles on experts, 
nor {\em ii)} relies solely on unreliable 
embeddings with missing modalities.
By comparing task embeddings from partial observations with their complete counterparts,
the router conditions expert specialization on representation deviation patterns, 
enabling adaptive handling of incomplete inputs.
\vspace{-3pt}
\section{Method}
\vspace{-3pt}
\label{sec:methods}
\sloppy

\begin{figure*}[t!]
  \centering
   \includegraphics[width=0.99\linewidth]{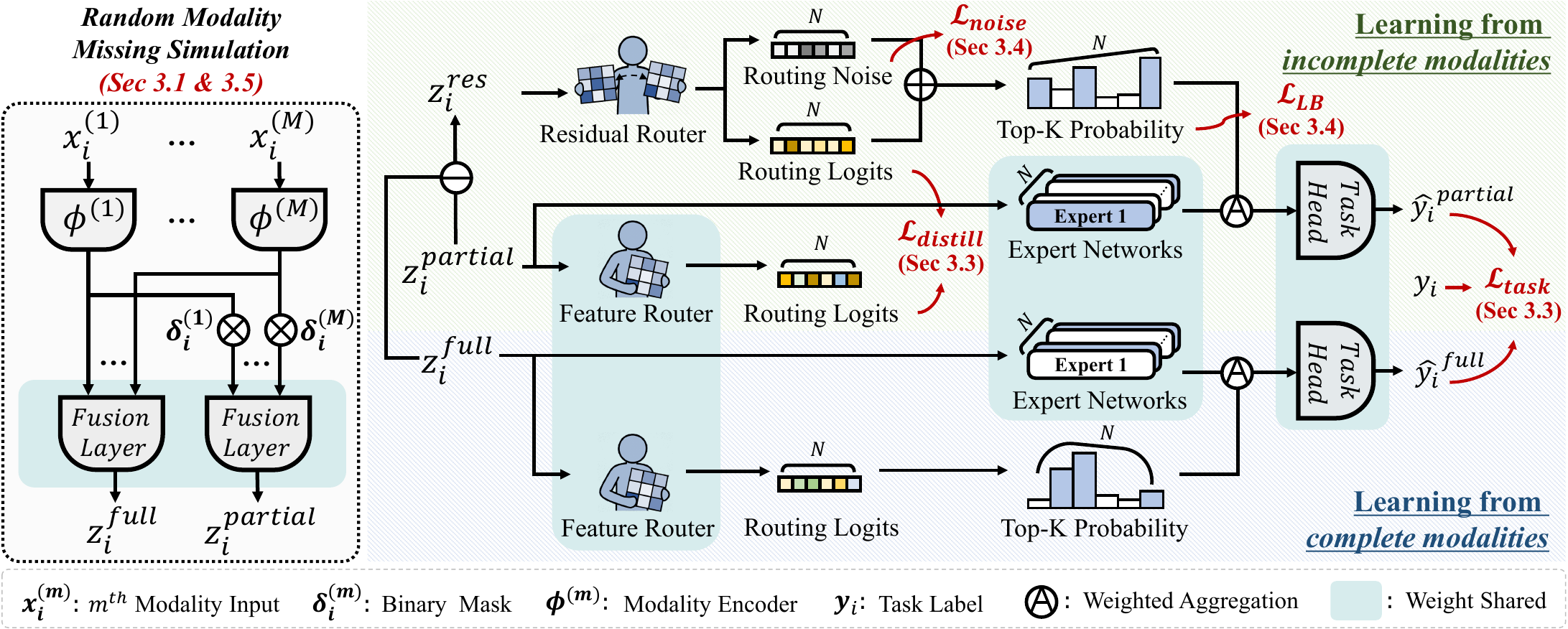}
    \vspace{-2mm}
   \caption{\small Overview of 
   MARS.
   During training, complete and incomplete features (\ie, $z_i^{full}$ and $z_i^{partial}$) are used to compute residuals that guide the residual router to specialize experts in distinct representation deviation patterns.
   The feature router learns to imitate this routing, enabling deployment without access to residuals.
   Discrepancy-aware noise and load-balancing further enhance expert robustness and diversity. 
   }
   \label{fig:overview}
    \vspace{-2mm}
\end{figure*}

The goal of MARS is {\em i)} to train experts to specialize in task-specific deviation caused by modality missingness, 
and {\em ii)} to ensure consistent utilization of this specialization during inference.
\cref{fig:overview} illustrates the overview, and following sections describe how its components and training procedure are designed.


\vspace{-3pt}
\subsection{Problem Definition}
\vspace{-3pt}
\label{sec:problem_definition}
Following recent works~\cite{DMRNet,MoMKE,SimMLM},
we consider the general setting of IML, 
where some modalities may be missing at inference while remaining fully available during training. 
Given a dataset, each sample 
is denoted as $x_i = \{x_i^{(m)}\}_{m=1}^{M}$, 
where $M$ indicates the number of modalities. 
Each sample with the $m$-th modality $x_i^{(m)}$ is encoded by a modality-specific encoder $\phi^{(m)}$, 
which yields $e_i^{(m)}$$=$$\phi^{(m)}(x_i^{(m)})$. 
To simulate missing-modality scenarios during training, 
each modality is stochastically deactivated using a binary indicator $\delta_i^{(m)} \in \{0,1\}$, 
where $\delta_i^{(m)} = 0$ masks its corresponding embedding $e_i^{(m)}$. 
Masked embeddings are concatenated and fused by a shared fusion layer $f_{\text{fuse}}$ to obtain a unified multimodal representation as
\begin{equation}
    z_i = f_{\text{fuse}}\!\left(
    \text{Concat}\big(\{\delta_i^{(m)}\,e_i^{(m)}\}_{m=1}^{M}\big)
    \right).
\end{equation}
This task representation $z_i$ is later used to compute task-specific predictions.

In previous methods, models process only one masked input (i.e., a single modality combination) per sample, 
from which a single task representation $z_i$ is extracted and used for prediction.
In contrast, 
MARS 
jointly operates on two fused features of the same sample,
one obtained from {\em the complete input} and the other from {\em the masked input}. 
By directly comparing these features,
MARS learns 
an explicit notion of 
how representations change under missing modalities.

\noindent\textbf{Preliminaries.}
A sparse Mixture-of-Experts model~\cite{sparseMoE} replicates some parts of the network as multiple expert modules. 
A router assigns scores to experts based on the input representation and aggregates the top-scoring expert outputs through weighted average.
The router’s decision encourages experts
to specialize across diverse input patterns with 
a modest increase in computational cost.

\subsection{MoE with Privileged Residual Router}
\label{sec:residual_router}
When all modalities are available, we refer to resulting task representation $z_i$ 
as a complete feature $z_i^\text{full}$, 
otherwise as an incomplete feature $z_i^\text{partial}$:
\begin{equation}
z_i =
\begin{cases}
z_i^{\text{full}}, & \text{if }~ \forall m,~\delta_i^{(m)} = 1 \\[1pt]
z_i^{\text{partial}}, & \text{otherwise}.
\end{cases}
\end{equation}
The MoE module is trained under a privileged setting 
where both $z_i^{\text{full}}$ and $z_i^{\text{partial}}$ are 
provided.
Their difference $z_i^\text{res}$ provides a residual signal that explicitly characterizes missingness-induced representation deviation patterns: 
\begin{equation}
    z_i^{\text{res}} = z_i^{\text{full}} - z_i^{\text{partial}}.
\end{equation}
This $z_i^{\text{res}}$ is fed into a {\em residual router} 
$\mathcal{R}_{\text{res}}$, 
which guides each expert to specialize in how the routed feature $z_i^\text{partial}$ deviates.
The $\mathcal{R}_{\text{res}}$ yields a clean logit vector $l_i^{\text{res}}\in\mathbb{R}^N$ 
and an estimated noise standard deviation $\sigma_i\in\mathbb{R}^N$, 
where $N$ denotes the number of experts.
To encourage diverse expert exploration,
we adopt {\em noisy top-$K$ routing}~\cite{sparseMoE} where
Gaussian noise $\epsilon$ is injected 
to make logits $l_i^{\text{res}}$ noisy as
\begin{equation}
\label{eq:noise}
\tilde{l}_i^{\text{res}} = l_i^{\text{res}} + \epsilon \sigma_i,
\quad \epsilon \sim \mathcal{N}(0, I).
\end{equation}
The top-$K$ experts with the highest noisy logits $\tilde{l}_i^{\text{res}}$ are selected 
to promote specialization while keeping the computation efficient.
Their routing probabilities $p_i^{\text{res}}$ are then computed using top-$K$ softmax operator $\text{Softmax}_K(\cdot)$, 
which applies the softmax over the selected top-$K$ logits while assigning zero to the rest:
\begin{equation}
p_i^{\text{res}} = \text{Softmax}_K(\tilde{l}_i^{\text{res}})\in\mathbb{R}^N.
\end{equation}
Let $(p^\text{res}_i)_j$ denote the routing probability of expert $j$ for sample $i$.
Each expert $E_j(\cdot)$ processes $z_i^{\text{partial}}$, 
and its output is weighted by $(p_i^{\text{res}})_j$. 
The aggregated output is then passed to a task head $h(\cdot)$, which produces a prediction $\hat{y}_i^\text{partial}$: 
\begin{equation}
\label{eq:y_partial}
\hat{y}_i^{\text{partial}}
= h\!\left(\sum_{j=1}^{N} (p_i^{\text{res}})_j \, E_j(z_i^{\text{partial}})\right).
\end{equation}
This residual routing directly supervises deviation-aware expert specialization.

{\em Remarks.} The formulation above 
faces two practical issues:  
1) $z_i^{\text{res}}$ is unavailable at inference when complete modalities are missing.
2) When all modalities are present, $z_i^{\text{res}}$ becomes zero, 
resulting in identical routing
across
samples. 
We therefore
introduce a {\em feature router} that operates without privileged information.


\vspace{-3pt}
\subsection{Preparing for Unprivileged Inference}
\vspace{-3pt}
\label{sec:feature_router}
To enable practical deployment 
without residual signal,
we introduce a feature router $\mathcal{R}_{\text{fea}}$ that 
performs routing using only the available embedding $z_i\in\{z_i^\text{full},z_i^\text{partial}\}$. 
It is designed 
as the deployable router at inference, 
capable of handling both complete and incomplete modality conditions.

Under full modality condition, 
residual $z_i^{\text{res}}$ inherently becomes zero.
Routing is thus entirely handled 
by $\mathcal{R}_{\text{fea}}$ as  $p_i^\text{fea}$$\,=\,$$\text{Softmax}_K(\mathcal{R}_{\text{fea}}(z_i^{\text{full}}))$. 
The task prediction $\hat{y}_i^\text{full}$ follows the same aggregation process in \cref{eq:y_partial} 
as $\hat{y}_i^{\text{full}}$
= $h\!\left(\sum_{j=1}^{N} (p^{\text{fea}}_i)_j E_j(z_i^{\text{full}})\right)$.
With a task-specific objective $\ell$ (e.g., cross-entropy) and target label $y_i$, 
the task-loss $\mathcal{L}_{task}$ combines both full and partial cases as
\begin{equation}
\mathcal{L}_{\text{task}} =
\ell(\hat{y}_i^{\text{full}}, y_i) +
\ell(\hat{y}_i^{\text{partial}}, y_i).
\end{equation}

For incomplete inputs, $\mathcal{R}_{\text{res}}$ performs privileged routing using $z_i^{\text{res}}$ to obtain $\hat{y}_i^{\text{partial}}$. 
In parallel, $\mathcal{R}_{\text{fea}}$ processes $z_i^{\text{partial}}$ but is optimized only 
through a distillation loss $\mathcal{L}_\text{distill}$ that aligns its logits $l_i^\text{fea}$ to those of $\mathcal{R}_{\text{res}}$ as
\begin{equation}
\mathcal{L}_{\text{distill}} =
D_{\mathrm{KL}}\!\left(
\text{Softmax}(\text{GradStop}(l_i^{\text{res}}))
\parallel
\text{Softmax}(l_i^{\text{fea}})
\right),
\end{equation}
where $D_{\mathrm{KL}}$ denotes the Kullback–Leibler divergence, and $\text{GradStop}(\cdot)$ detaches gradients to prevent updates. 
This enables $\mathcal{R}_{\text{fea}}$ to learn from 
the privileged $\mathcal{R}_{\text{res}}$, 
making it the sole deployable router with similar expert selection at inference.

\vspace{-3pt}
\subsection{Discrepancy-Aware Noise Regularization}
\vspace{-3pt}
\label{sec:noise}
With noisy routing (\cref{sec:residual_router}), 
MoE frameworks typically incorporate 
a load–importance balancing loss $\mathcal{L}_{\text{LB}}$~\cite{sparseMoE} 
for 
balanced expert utilization. 
Let $B$ denote the mini-batch size, and
define the importance and load of expert $j$ as
\begin{equation}
\text{importance}_j = \sum_{i=1}^{B} (p^\text{res}_i)_j, 
\quad
\text{load}_j = \sum_{i=1}^{B} \mathbb{I}[(p^\text{res}_i)_j > 0],
\end{equation}
where $\mathbb{I}(\cdot)$ is an indicator function that equals 1 if the condition holds, and 0 otherwise.
The load-importance balancing loss $\mathcal{L}_{\text{LB}}$ is then computed as
\begin{equation}\small
\mathcal{L}_{\text{LB}} =
\text{CV}^2\!\left(\{\text{importance}_j\}_{j=1}^N\right)
+ \text{CV}^2\!\left(\{\text{load}_j\}_{j=1}^N\right),
\end{equation}
where squared coefficient of variation $\text{CV}^2(x)=(\sigma(x)/\mu(x))^2$ measures the uniformity across experts. 
This $\mathcal{L}_\text{LB}$
prevents expert collapse and ensures balanced participation, 
while the injected routing noise $\epsilon\sigma_i$ (\cref{eq:noise}) promotes stochastic expert exploration. 
Beyond this conventional role, we 
further leverage the noise to
mitigate the unavoidable
routing discrepancy between training and inference.

The residual router $\mathcal{R}_{\text{res}}$ provides an optimal supervision 
by leveraging both $z_i^\text{partial}$ and $z_i^\text{full}$. 
However, despite the distillation loss $\mathcal{L}_{\text{distill}}$, 
$\mathcal{R}_{\text{fea}}$ cannot fully reproduce the behavior of $\mathcal{R}_{\text{res}}$
since it operates only with $z_i^\text{partial}$. 
This inevitably widens the train–test routing gap which causes inconsistent expert assignments. 

To mitigate this gap, we introduce a discrepancy-aware noise regularization $\mathcal{L}_{\text{noise}}$, 
which scales the noise variance $(\sigma_i^2)_j$ with the routing discrepancy of expert $j$. 
Higher noise variance induces the stochasticity of expert selection, 
allowing experts to remain robust even when the $\mathcal{R}_{\text{fea}}$ replaces $\mathcal{R}_{\text{res}}$ at inference.

Let $l_i^{\text{res}}$ and $l_i^{\text{fea}}$ denote the clean logits from $\mathcal{R}_{\text{res}}$ and $\mathcal{R}_{\text{fea}}$, respectively.
We extract the top-$K$ expert indices $\mathcal{T}_i^{\text{res}},\mathcal{T}_i^{\text{fea}}\in\mathbb{R}^K$ from each router as
\begin{equation}
\mathcal{T}_i^{\text{res}} = \text{TopK}(l_i^{\text{res}}, K), \quad
\mathcal{T}_i^{\text{fea}} = \text{TopK}(l_i^{\text{fea}}, K),
\end{equation}
and define their union $\mathcal{U}_i$$=$$\mathcal{T}_i^{\text{res}}$$\,\cup\,$$\mathcal{T}_i^{\text{fea}}$. 
For each $j\,$$\in$$\,\mathcal{U}_i$, we measure the routing discrepancy $(m_i)_j$ 
as the absolute difference between the normalized clean logits:
\begin{equation}
(m_i)_j = 
\big|
\text{Softmax}(l_i^{\text{res}})_j -
\text{Softmax}(l_i^{\text{fea}})_j
\big|,
\quad j \in \mathcal{U}_i.
\end{equation}
We then sort experts in $\mathcal{U}_i$
in descending order of $(m_i)_j$,
which yields a permutation $\pi_i$ such that
$(m_i)_{\pi_i(1)}$$\ge$$
\dots$$ \ge$$
(m_i)_{\pi_i(|\mathcal{U}_i|)}$.
To promote larger noise for experts with greater discrepancy, 
we penalize cases where the ordering of $(\sigma_i^2)_j$ contradicts that of $(m_i)_j$. 
Thus, the noise regularization $\mathcal{L}_{\text{noise}}$ is formulated as
\begin{equation}\small
\mathcal{L}_{\text{noise}}
=
\frac{1}{|\mathcal{U}_i| - 1}
\sum_{j=1}^{|\mathcal{U}_i| - 1}
\mathrm{Softplus}\Big(
\log(\sigma_i)_{\pi_i(j+1)}^2
-
\log(\sigma_i)_{\pi_i(j)}^2
\Big),
\label{eq:lvar}
\end{equation}
which softly enforces
$(\sigma_i^2)_{\pi_i(1)} \ge \dots \ge (\sigma_i^2)_{\pi_i(|\mathcal{U}_i|)}$. 
As experts with greater discrepancy $(m_i)_j$ are guided to exhibit higher routing noise $(\sigma_i^2)_j$, 
this encourages exploration of alternative experts where the routers 
disagree.
By being stochastically activated,
these experts become more robust to test-time routing mismatch.

Finally, the overall training objective $\mathcal{L}_{\text{total}}$ is defined as
\begin{equation}
\small
\mathcal{L}_{\text{total}}
=
\lambda_{\text{task}}\mathcal{L}_{\text{task}}
+ \lambda_{\text{LB}}\mathcal{L}_{\text{LB}}
+ \lambda_{\text{distill}}\mathcal{L}_{\text{distill}}
+ \lambda_{\text{noise}}\mathcal{L}_{\text{noise}},
\end{equation}
where each $\lambda$ controls the balance between individual loss terms $\mathcal L$. 

\vspace{-3pt}
\subsection{Top-$K$ Discrepancy-Guided Modality Sampling}
\vspace{-3pt}
\label{sec:sampling}
We further introduce an adaptive sampling strategy to balance the learning 
progress across different incomplete modality configurations. 
At the beginning of training, configurations are sampled uniformly,
while the full-modality case is always included to provide complete supervision.
After several warm-up epochs, the sampling probabilities are updated
according to the routing discrepancy. 

Let $\mathcal{C} = \{c_1, \ldots, c_{|\mathcal{C}|}\}$ denote all feasible incomplete modality 
combinations, where each $c_{j\in\{1,\ldots,|\mathcal{C}|\}}$ is represented by a binary vector 
composed of the masking indicators 
$\{\delta^{(m)} \in \{0,1\}\}_{m=1}^{M}$.
We denote $q_j$ as the sampling probability of configuration $c_j$ satisfying 
$\sum_{j=1}^{|C|} q_j = 1$. 
For each training epoch, we measure the average top-$K$ routing disagreement $d_j$ between the routers
over all samples whose masked configuration corresponds to $c_j$ as
\begin{equation}
d_j = 1 -
\frac{1}{|\mathcal{D}_{c_j}|}
\sum_{i \in \mathcal{D}_{c_j}}
\frac{|\mathcal{T}_i^{\text{res}} \cap \mathcal{T}_i^{\text{fea}}|}{K},
\end{equation}
where $\mathcal{D}_{c_j}$ denotes the set of training samples assigned to $c_j$
throughout the current epoch.
At the end of the $t$-th iteration, discrepancies are normalized by a 
softmax to update the next sampling probabilities $q_j^{(t+1)}$ as
\begin{equation}
\label{eq:sampling_prob}
q_j^{(t+1)} =
\frac{\exp(d_j / \tau)}
{\sum_{k=1}^{|\mathcal{C}|} \exp(d_k / \tau)},
\end{equation}
where the temperature $\tau$ adjusts the smoothness. 
Combinations showing higher router disagreement $d_j$ are sampled more 
frequently in subsequent epochs, allowing the model to focus on modality settings 
that are harder to align.

\vspace{-3pt}
\section{Experiments} 
\vspace{-3pt}
We evaluate our method on 1) multimodal spoof classification, 2) material segmentation, 3) emotion recognition, and 4) food classification using CASIA-SURF~\cite{CASIASURF}, MCubeS~\cite{MCubeS}, CREMA-D~\cite{crema}, and UPMC Food-101~\cite{food101}.
For all datasets, missing-modality conditions follow the common protocols (\cref{sec:problem_definition}) used in previous IML works~\cite{MMANet,DMRNet,MoMKE,SimMLM}.
Due to space limitations, detailed dataset descriptions and result analyses for CREMA-D~\cite{crema} (\cref{tab:cremad_result}) and UPMC Food-101~\cite{food101} (\cref{tab:food_result}) are provided in \cref{sec:additional_experiment}.

\subsection{Datasets}
\vspace{-3pt}
\label{sec:dataset}
\textbf{CASIA-SURF}~\cite{CASIASURF} is a large-scale multimodal face anti-spoofing benchmark that contains RGB, Depth, and Infrared (IR) modalities. 
Each modality provides complementary cues for distinguishing live faces from spoofing attacks. 
The dataset offers binary labels indicating live or spoof, 
with $N$$=$29k, 1k, and 57k samples for training, validation, and testing data.
Performance is evaluated using the Average Classification Error Rate (ACER, \%)~\cite{CASIASURF}, 
which represents the percentage of misclassified samples.
CASIA-SURF has been adopted as a standard benchmark for incomplete multimodal classification~\cite{MMANet,DMRNet}.

\noindent\textbf{MCubeS}~\cite{MCubeS} is a multimodal material segmentation dataset composed of 500 samples captured from 42 diverse outdoor scenes. 
Each sample includes four modalities: RGB, Degree of Linear Polarization (DoLP), Angle of Linear Polarization (AoLP), and Near-Infrared (NIR) reflectance. 
Every pixel is annotated with one of 20 material categories such as {\em concrete}, {\em water}, and {\em metal}. 
The official split includes 302 training, 96 validation, and 102 test samples. 
Performance is measured by the mean Intersection-over-Union (mIoU), 
where higher values indicate better segmentation accuracy.
We employ this recent dataset in IML scenarios
to assess segmentation robustness under missing-modality conditions. 


\noindent\textbf{CREMA-D}~\cite{crema} and \textbf{UPMC Food-101}~\cite{food101} are multimodal classification datasets comprising {\em audio–visual} and {\em image–text} modalities, respectively.
They enable evaluation beyond vision-centric datasets.
We follow the experimental setups of DMRNet~\cite{DMRNet} for CREMA-D and SimMLM~\cite{SimMLM} for UPMC Food-101.

\vspace{-3pt}
\subsection{Implementation Details}\label{sec:implementary}
\vspace{-3pt}
For CASIA-SURF~\cite{CASIASURF} and MCubeS~\cite{MCubeS}, 
we adopt a sparse MoE~\cite{sparseMoE} framework with 16 experts, 
where the top five are activated per sample.
The residual router $\mathcal{R}_{res}$ employs a single linear projection layer,
and the feature router $\mathcal{R}_{fea}$ consists of a two-layer MLP.
All experiments are run on a single NVIDIA RTX 6000 ADA.

\noindent\textbf{CASIA-SURF.} 
Following DMRNet~\cite{DMRNet}, we employ a ResNet-18~\cite{resnet} backbone with 
a binary indicator $\delta^{(m)}$ 
(\cref{sec:problem_definition}). 
The final convolution block is used as the expert set. 
Training is performed for 100 epochs using SGD~\cite{sgd} with a learning rate of $5$$\times$$10^{-4}$, 
weight decay of $5$$\times$$10^{-4}$, and a batch size of 64. 
Loss weights are $\lambda_\mathrm{task}$$=$$1$, $\lambda_\mathrm{LB}$$=$$0.05$, $\lambda_\mathrm{distill}$$=$$1$, and $\lambda_\mathrm{noise}$$=$$0.01$. 
Sampling begins at epoch 20. 

\noindent\textbf{MCubeS.} 
We build on DeepLab v3+~\cite{deeplab} with the same Bernoulli indicator scheme. 
The final layer before the task head serves as the expert set. 
Models are trained for 500 epochs using SGD with a learning rate of $5$$\times$$10^{-3}$, weight decay of $5$$\times$$10^{-4}$, and a batch size of 8. 
Loss weights are set to $\lambda_\mathrm{task}$$=$$1$, $\lambda_\mathrm{LB}$$=$$0.01$, $\lambda_\mathrm{distill}$$=$$0.05$, and $\lambda_\mathrm{noise}$$=$$0.1$, and our sampling strategy starts from epoch 30.

\noindent\textbf{CREMA-D} and \textbf{UPMC Food-101.} Refer to \cref{sec:additional_experiment}.


\vspace{-3pt}
\subsection{Baselines}
\vspace{-3pt}
We compare our method with recent state-of-the-art approaches. 
For classification, we report the DMRNet~\cite{DMRNet} results from the original paper 
and re-implement Flex-MoE~\cite{FlexMoE}, MoMKE~\cite{MoMKE}, and SimMLM~\cite{SimMLM} using a ResNet-18~\cite{resnet} backbone for fair comparison. 
Since the original Flex-MoE assumes incomplete modalities during training, we adapt it by providing both complete and randomly dropped modalities via $\delta^{(m)}$. 
For segmentation, all baselines are re-implemented on DeepLab~v3+~\cite{deeplab}. 
All hyperparameters are tuned for optimal performance. 
Further experimental details are given in \cref{sec:details}. 

\begin{table*}[!t]
\centering
\caption{\small Classification results (ACER ↓) on CASIA-SURF~\cite{CASIASURF} under all modality combinations.
MARS consistently outperforms baselines and achieves the lowest error rate across all settings, demonstrating strong robustness to missing modalities.
The best and second-best results are highlighted with \textbf{bold} and \underline{underlined}, respectively.
(\CIRCLE: Presence, \Circle: Absence)
}
\vspace{-2mm}
\renewcommand{\arraystretch}{0.8}
\resizebox{\textwidth}{!}{%
\begin{tabular}{c c c | c c c c c c c c c c c c}
\toprule
\multicolumn{3}{c|}{Modality} & \multicolumn{12}{c}{Methods} \\
\midrule

RGB & Depth & IR & ResNet-18 & HeMIS & LCR & RFNet & mmFormer & ShaSpec & MMANet & DMRNet & Flex-MoE & MoMKE & SimMLM & \cellcolor{gray!15}MARS\\
\midrule
\CIRCLE & \Circle & \Circle & 11.75 & 14.36 & 13.44 & 12.43 & 11.15 & 11.57 & 8.57 & 8.23 & 9.10 & 10.04 & \underline{7.82} & \cellcolor{gray!15}\textbf{6.92} \\
\Circle & \CIRCLE & \Circle & 5.87 & 4.70 & 4.40 & 4.17 & 3.67 & 6.25 & 2.27 & 2.01 & 2.63 & \underline{1.95} & 2.42 & \cellcolor{gray!15}\textbf{1.87} \\
\Circle & \Circle & \CIRCLE & 16.62 & 16.21 & 15.26 & 14.69 & 13.99 & 10.71 & 10.04 & 8.98 & \underline{7.25} & 12.57 & 9.05 & \cellcolor{gray!15}\textbf{3.96} \\
\midrule
\CIRCLE & \CIRCLE & \Circle & 4.61 & 3.23 & 3.32 & 2.23 & 1.93 & 3.11 & 1.61 & 1.21 & 1.27 & 1.20 & \underline{1.13} & \cellcolor{gray!15}\textbf{1.06} \\
\CIRCLE & \Circle & \CIRCLE & 6.68 & 6.27 & 5.16 & 4.27 & 4.77 & 4.23 & 3.01 & \underline{3.00} & 3.58 & 5.09 & 3.38 & \cellcolor{gray!15}\textbf{2.09} \\
\Circle & \CIRCLE & \CIRCLE & 4.95 & 3.68 & 3.53 & 3.22 & 3.10 & 2.52 & 1.18 & \underline{0.80} & 1.19 & 1.06 & 1.15 & \cellcolor{gray!15}\textbf{0.66} \\
\midrule
\CIRCLE & \CIRCLE & \CIRCLE & 2.21 & 1.97 & 1.88 & 1.18 & 1.94 & 1.79 & 0.87 & 0.66 & 0.84 & 0.79 & \underline{0.63} & \cellcolor{gray!15}\textbf{0.45} \\
\midrule
\multicolumn{3}{c|}{Average} & 7.52 & 7.18 & 6.71 & 6.02 & 5.93 & 5.59 & 3.94 & \underline{3.58} & 3.69 & 4.67 & 3.66 & \cellcolor{gray!15}\textbf{2.43}\\
\bottomrule
\end{tabular}%
}
\vspace{-7.5mm}
\label{tab:classification}
\end{table*}

\vspace{-5pt}
\subsection{Spoof Classification Results}
\vspace{-3pt}
\label{sec:clf_result}
\cref{tab:classification} summarizes the spoof classification results on the CASIA-SURF dataset.  
Among the baselines, DMRNet~\cite{DMRNet} achieves the lowest average ACER of 3.58, 
while 
MARS 
achieves 2.43 reducing it by 1.15. 
The model consistently attains the lowest ACER across all modality combinations, 
demonstrating strong robustness to missing modalities. 
The most challenging case occurs when only the IR 
is available. 
In this setting, 
MARS
reduces the ACER from 8.98 (in DMRNet) to 3.96, 
effectively reducing the error by more than half. 
These results confirm that MARS robustly handles representation deviations induced by modality removal.

\begin{figure}[t]
\centering
\Large
\begin{minipage}[t]{0.555\linewidth}
\centering
\captionof{table}{\small Segmentation results (mIoU $\uparrow$) on MCubeS~\cite{MCubeS} with all RGB combinations.}
\label{tab:segmentation}
\renewcommand{\arraystretch}{1.07}
\resizebox{\linewidth}{!}{%
\begin{tabular}{c c c c | c c c c c >{\columncolor{gray!15}}c}
\toprule
\rotatebox{12}{RGB} & \rotatebox{10}{AoLP} & \rotatebox{12}{DoLP} & \rotatebox{12}{NIR} & \rotatebox{12}{\makecell{DeepLab\\v3+}} & \rotatebox{12}{DMRNet} & \rotatebox{12}{\makecell{Flex-\\MoE}} & \rotatebox{12}{MoMKE} & \rotatebox{12}{SimMLM} & \rotatebox{12}{MARS} \\
\midrule
\CIRCLE & \Circle & \Circle & \Circle & 0.4249 & \underline{0.4647} & 0.4490  & 0.4150 & 0.4462 & \textbf{0.4730} \\
\midrule
\CIRCLE & \CIRCLE & \Circle & \Circle & 0.4238 & \underline{0.4682} & 0.4528 & 0.4241 & 0.4473 & \textbf{0.4767} \\
\CIRCLE & \Circle & \CIRCLE & \Circle & 0.4249 & \underline{0.4670} & 0.4529 & 0.4326 & 0.4420 & \textbf{0.4753} \\
\CIRCLE & \Circle & \Circle & \CIRCLE & 0.4247 & \underline{0.4672} & 0.4515 & 0.4008 & 0.4355 & \textbf{0.4769} \\
\midrule
\CIRCLE & \CIRCLE & \CIRCLE & \Circle & 0.4269 & \underline{0.4694} & 0.4559 & 0.4173 & 0.4427 & \textbf{0.4777} \\
\CIRCLE & \CIRCLE & \Circle & \CIRCLE & 0.4269 & \underline{0.4701} & 0.4554 & 0.4325 & 0.4286 & \textbf{0.4808} \\
\CIRCLE & \Circle & \CIRCLE & \CIRCLE & 0.4261 & \underline{0.4691} & 0.4558 & 0.4238 & 0.4224 & \textbf{0.4776} \\
\midrule
\CIRCLE & \CIRCLE & \CIRCLE & \CIRCLE & 0.4271 & \underline{0.4712} & 0.4586 & 0.4376 & 0.4220 & \textbf{0.4808} \\
\midrule
\multicolumn{4}{c|}{\textbf{Average}} & 0.4257 & \underline{0.4683} & 0.4540 & 0.4273 & 0.4367 & \textbf{0.4773} \\
\bottomrule
\end{tabular}}
\end{minipage}
\hfill
\begin{minipage}[t]{0.43\linewidth}
\centering
\captionof{table}{\small Results on CREMA-D~\cite{crema}.}
\label{tab:cremad_result}
\renewcommand{\arraystretch}{0.85}
\resizebox{\linewidth}{!}{%
\begin{tabular}{c c | c c c c}
\toprule
Audio & Visual & ShaSpec & MMANet & DMRNet & \cellcolor{gray!15}MARS \\
\midrule
\CIRCLE & \Circle & \underline{59.86} & 58.89 & 58.87 & \cellcolor{gray!15}\textbf{61.04} \\
\Circle & \CIRCLE & 47.17 & 47.31 & \underline{55.10} & \cellcolor{gray!15}\textbf{59.43} \\
\midrule
\CIRCLE & \CIRCLE & 66.80 & 67.87 & \underline{70.10} & \cellcolor{gray!15}\textbf{76.10} \\
\midrule
\multicolumn{2}{c|}{Average} & 57.94 & 57.66 & \underline{61.35} & \cellcolor{gray!15}\textbf{65.52} \\
\bottomrule
\end{tabular}
}
\vspace{-3mm}
\renewcommand{\arraystretch}{0.845}
\captionof{table}{\small Results on Food-101~\cite{food101}.}
\label{tab:food_result}
\resizebox{\linewidth}{!}{%
\begin{tabular}{cc|cccc}
\toprule
 Image & Text 
& ShaSpec
& MoMKE
& SimMLM
& \cellcolor{gray!15}MARS \\
\midrule
 \CIRCLE & \Circle
& 69.22 & 70.46 & \underline{72.20} & \cellcolor{gray!15}\textbf{91.86} \\
 \Circle & \CIRCLE 
& 86.55 & 86.59 & \underline{87.20} & \cellcolor{gray!15}\textbf{90.18} \\
\midrule
 \CIRCLE & \CIRCLE 
& \underline{92.73} & 92.71 & \textbf{94.99} & \cellcolor{gray!15}92.72 \\
\midrule
 \multicolumn{2}{c|}{Average}
& 82.83 & 83.25 &  \underline{84.81} & \cellcolor{gray!15}\textbf{91.59} \\
\bottomrule
\end{tabular}
}
\end{minipage}
\vspace{-15pt}
\end{figure}

\vspace{-3pt}
\subsection{Material Segmentation Results}
\vspace{-3pt}
\cref{tab:segmentation} reports the segmentation results on the MCubeS~\cite{MCubeS} dataset.
Following the initial work \cite{MCubeS}, 
we focus on combinations that include the RGB, 
since non-RGB modalities, such as NIR, AoLP, or DoLP, alone lack sufficient radiometric diversity to discern all classes. 
Consistent with the classification results, 
DMRNet~\cite{DMRNet} achieves the best baseline performance with an average mIoU of 0.4683.
MARS improves it to 0.4773, surpassing baselines across all combinations. 

\begin{figure*}[!t]
  \centering
   \includegraphics[width=0.99\linewidth]{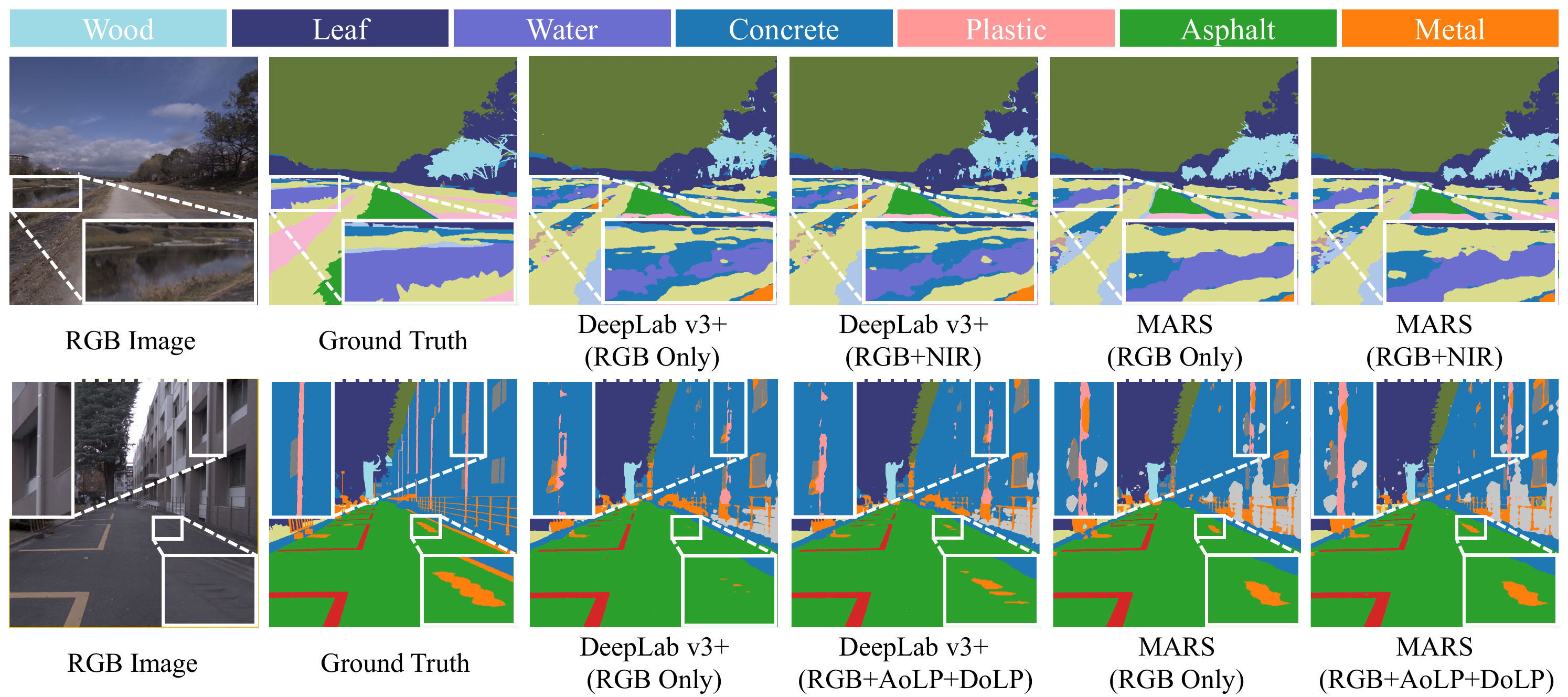}
  \vspace{-3mm}
   \caption{ \small 
   Qualitative results on multimodal material segmentation (MCubeS~\cite{MCubeS}).
    Top: When near-infrared (NIR) information is missing, DeepLab~v3+~\cite{deeplab} degrades in recognizing water and reflected objects, whereas MARS remains relatively consistent.
    Bottom: In the absence of polarization cues (AoLP/DoLP), MARS correctly separates metallic and dielectric materials even under incomplete inputs.}
   \label{fig:qualitative}
   \vspace{-4mm}
\end{figure*}

\cref{fig:qualitative} presents qualitative comparisons on test samples from the MCubeS dataset.
When compared to DeepLab~v3+ \cite{deeplab}, 
MARS 
provides clearer and more consistent segmentation results.
When the NIR 
is missing, which helps identify water, 
DeepLab~v3+ often misinterprets reflected surfaces over water as separate objects, while 
MARS 
remains robust to such misleading cues. 
Similarly, polarization cues such as AoLP and DoLP are important for distinguishing metallic and dielectric materials.
Without 
them, 
the baseline predictions frequently confuse concrete, metal, and plastic, 
whereas 
MARS 
produces more stable segmentation results.
These observations suggest that 
MARS 
better handles regions that are ambiguous under the given modalities, 
as {\em residual guidance encourages experts to focus on reliable cues} rather than misleading artifacts.

\vspace{-4pt}
\section{Model Behavior Analysis}
\vspace{-3pt}
We analyze the MARS framework to better understand the effect of its key components and the behavior of expert routing under missing-modality scenarios.

\vspace{-5pt}
\subsection{Ablation Study}
\vspace{-5pt}
\cref{tab:ablation} presents the ablation results on CASIA-SURF and MCubeS. 
Starting from each baseline, ResNet-18 and DeepLab~v3+, 
the introduction of the MoE structure improves performance by diversifying strategies on inputs. 
Adding residual routing further enhances those MoE by leveraging the representational deviations for expert specialization.
Variance ordering brings an additional gain by guiding the routing noise to account for train-test gap induced from router change. 
Finally, discrepancy sampling, which selectively emphasizes large gap between routers cases during training, yields the best overall performance. 

We add an experiment to solidify our key idea that {\em residual signal promotes optimal expert specializations}.
The Oracle routing experiment assumes access to full modalities even at test time, 
where the true residuals are fed to the residual router for inference. 
This setting achieves near-perfect classification accuracy (ACER 0.94), 
confirming that the experts themselves are already well specialized to modality missingness. 
Interestingly, in segmentation, our model slightly surpasses the Oracle (mIoU 0.4773 vs. 0.4768). 
We attribute this to the noise regularization, which 
prevents overreliance on the residual routing and 
encourages the experts to learn complementary decision boundaries for the feature routing. 

\begin{table*}[t]
\centering

\begin{minipage}[t]{0.46\linewidth}
\centering
\captionof{table}{\small Ablation study. Performance is averaged over all modality configurations. ``+'' denotes cumulative inclusion.}
\label{tab:ablation}
\vspace{-2mm}
\renewcommand{\arraystretch}{1.125}

\resizebox{\linewidth}{!}{%
\begin{tabular}{lcc}
\toprule
Method & \multicolumn{1}{c}{\makecell{CASIA-SURF \\ (ACER $\downarrow$)}} 
       & \multicolumn{1}{c}{\makecell{MCubeS \\ (mIoU $\uparrow$)}} \\
\midrule
ResNet-18 / DeepLab~v3+ & 7.52 & 0.4257 \\
+ MoE & 4.12 & 0.4327 \\
+ Residual Routing & 3.78 & 0.4661 \\
+ Noise Regularization & 2.72 & 0.4750 \\
\rowcolor{gray!15}+ Discrepancy-guided sampling & 2.43 & \textbf{0.4773} \\
\midrule
Oracle Routing & \textbf{0.94} & 0.4768 \\
\bottomrule
\end{tabular}%
}
\end{minipage}
\hfill
\begin{minipage}[t]{0.53\linewidth}
\centering
\captionof{table}{\small Error rate (ACER, \%) under expert deactivation on CASIA-SURF. Changes in ().}
\vspace{-3mm}
\renewcommand{\arraystretch}{0.875}
\resizebox{\linewidth}{!}{%
\begin{tabular}{c c c | c | c c c}
\toprule
\multicolumn{3}{c|}{Modality} & \multicolumn{4}{c}{Deactivated Expert} \\
\midrule
RGB & Depth & IR & None & \cellcolor{RawSienna!15}E12 & \cellcolor{GreenYellow!35}E0 & \cellcolor{Fuchsia!15}E5 \\
\midrule
\cellcolor{RawSienna!15}\CIRCLE & \cellcolor{RawSienna!15}\Circle & \cellcolor{RawSienna!15}\Circle & 6.92 & {\bf \cellcolor{RawSienna!15}7.46 (0.54$\uparrow$)} & 7.54 (0.62$\uparrow$) & 6.92 (0.00--)\\
\Circle & \CIRCLE & \Circle & 1.87 & 1.87 (0.00--) & 1.74 (0.13$\downarrow$) & 1.87 (0.00--) \\
\cellcolor{GreenYellow!35}\Circle & \cellcolor{GreenYellow!35}\Circle & \cellcolor{GreenYellow!35}\CIRCLE & 3.96 & 3.96 (0.00--) & {\bf \cellcolor{GreenYellow!35}7.58 (3.62$\uparrow$)} & 3.96 (0.00--) \\
\midrule
\CIRCLE & \CIRCLE & \Circle & 1.06 & 0.99 (0.07$\downarrow$) & 0.93 (0.13$\downarrow$) & 1.06 (0.00--) \\
\CIRCLE & \Circle & \CIRCLE & 2.09 & 1.52 (0.57$\downarrow$) & 1.73 (0.36$\downarrow$) & 2.09 (0.00--) \\
\Circle & \CIRCLE & \CIRCLE & 0.66 & 0.66 (0.00--) & 0.48 (0.18$\downarrow$) & 0.66 (0.00--) \\
\midrule
\cellcolor{Fuchsia!15}\CIRCLE & \cellcolor{Fuchsia!15}\CIRCLE & \cellcolor{Fuchsia!15}\CIRCLE & 0.45 & 0.31 (0.14$\downarrow$) & 0.36 (0.09$\downarrow$) & \cellcolor{Fuchsia!15} 0.93 (0.48$\uparrow$) \\
\midrule
\multicolumn{3}{c|}{Average} & 2.43 & 2.40 (0.03$\downarrow$) & 2.91 (0.48$\uparrow$) & 2.50 (0.07$\uparrow$) \\
\bottomrule
\end{tabular}%
}
\label{tab:expert_deactivate}
\end{minipage}
\end{table*}

\begin{figure}[!t]
  \centering
  \vspace{-3mm}
\includegraphics[width=0.99\columnwidth]{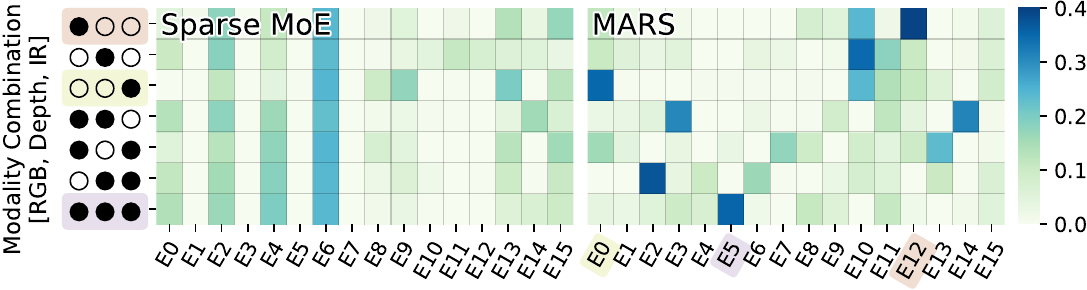}
    \vspace{-4mm}
   \caption{\small Top-$K$ routing probabilities of the feature router at inference on CASIA-SURF~\cite{CASIASURF}.
    Probabilities are averaged over modality combination.
    Our method shows diverse expert utilization, unlike the baseline MoE~\cite{sparseMoE} activating similar experts.}
   \label{fig:routing}
   \vspace{-5mm}
\end{figure}

\vspace{-3pt}
\subsection{Effectiveness of Residual Routing}
\vspace{-3pt}

\begin{figure*}[!t]
  \centering
    \includegraphics[width=\linewidth]{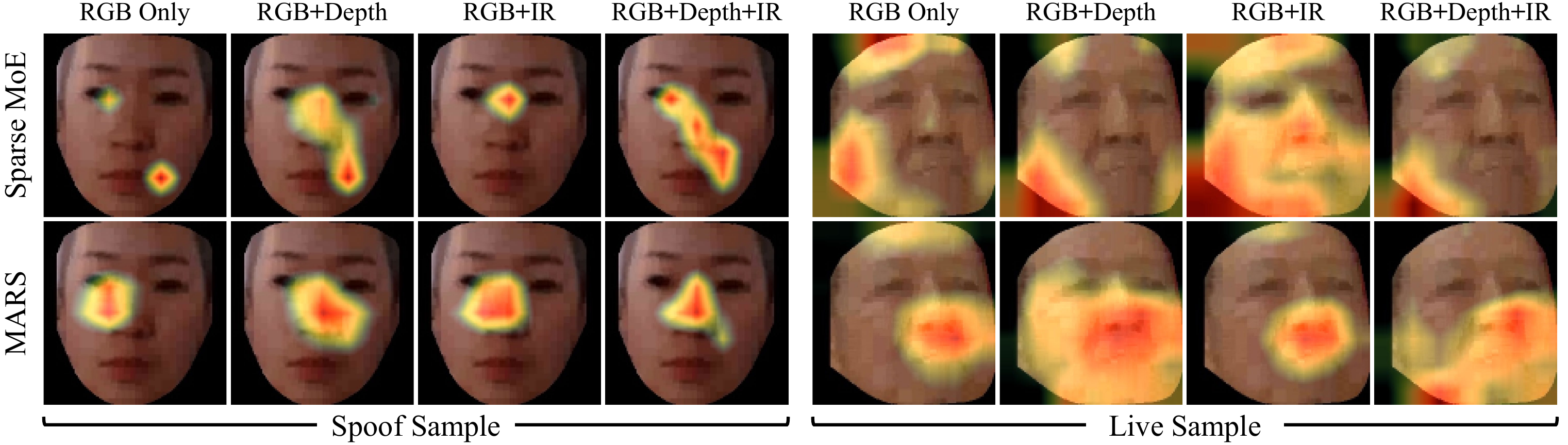}
    \vspace{-7mm}
   \caption{\small Grad-CAM~\cite{gradcam} visualization on CASIA-SURF~\cite{CASIASURF}.
    Heatmaps are generated by backpropagating the predicted logit of the target class through 
    $z_i$ and then overlaid on the RGB image. 
    Compared to the baseline MoE~\cite{sparseMoE}, which inconsistently attends to superficial artifacts (\eg, around the mouth corner), MARS produces stable and semantically meaningful activations (\eg, around the nose) across diverse combinations.}
   \label{fig:gradcam}
   \vspace{-9pt}
\end{figure*}

To show the effectiveness of residual routing in expert specialization,
we analyze the routing distribution (\cref{fig:routing}) and visualizations of feature attribution (\cref{fig:gradcam}).

\noindent\textbf{Routing Distribution.}
\cref{fig:routing} shows the routing probabilities of the feature router at test time on CASIA-SURF~\cite{CASIASURF}.
For each modality combination, we collect the top-$K$ routing probabilities and average them across all samples to depict the expert activation distribution.
To ensure a fair comparison, both the original MoE and our model share identical weight of hyperparameters including load-balancing weight $\lambda_\mathrm{LB}$.
In the baseline MoE, routing remains biased toward a few dominant experts irrespective of the modality combination.
In contrast, our residual routing exhibits distinct yet structured expert utilization among different combinations.
Most combinations activate at most two dominant experts, 
which rarely overlap to other combinations.
Especially, we identify strong associations 
between specific modality combinations and experts
(e.g., RGB–E12, IR–E0, Full–E5).
This implies that the router effectively operates on missingness.

To further verify whether the experts act responsible for the combination,
we deactivate certain experts at inference and summarize the corresponding performance drops for each combination in \cref{tab:expert_deactivate}.
The combinations most strongly associated with the expert consistently incur the largest error increase.
This confirms that experts with strong affinity to particular modality combinations
learn disentangled specializations aligned with distinct missing-modality patterns.

\noindent\textbf{Feature Attribution.}
\cref{fig:gradcam} visualizes Grad-CAM~\cite{gradcam} results on CASIA-SURF~\cite{CASIASURF}.
For test samples $x_i$, we 
backpropagate the predicted target logit $\hat{y_i}$ through the fused embedding $z_i$, 
and overlay the activation map on its RGB image $x_i^\text{RGB}$.
Each selected configuration includes one spoof and one live sample.

In the baseline MoE, attention regions vary largely across modality combinations, 
often highlighting superficial artifacts (e.g., mouth corner or forehead in RGB-only input).
In contrast, our method consistently focuses on stable facial regions, particularly around the nose, which serves as a key discriminative cue for spoof detection. 
It also adapts its attention to secondary areas, such as the eyes or cheeks, depending on the available signals.
This coherent attention pattern indicates that residual routing encourages reasoning based on reliable features
rather than overfitting to spurious modality-specific artifacts.


\begin{figure}[!t]
\centering
\begin{minipage}[t]{0.31\columnwidth}
  \centering
  \includegraphics[width=0.99\linewidth, height = 2.9cm]{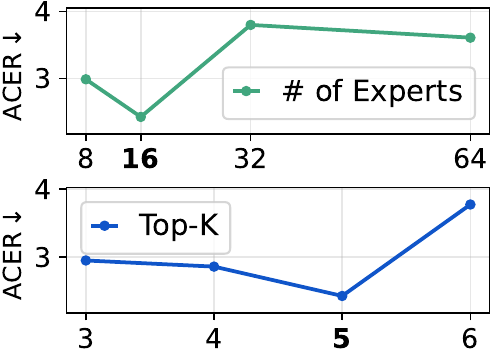}
    \vspace{-4.2mm}
    \subcaption{Sensitivity analysis.}
  \label{fig:sensitivity}
\end{minipage}
\hfill
\begin{minipage}[t]{0.37\columnwidth}
  \centering
  \vspace{-28.7mm}  
  \renewcommand{\arraystretch}{1.01}
  \resizebox{\linewidth}{!}{%
  \begin{tabular}{l|c|c}
    \toprule
    \textbf{Method} & \textbf{\#Params} & \textbf{FLOPs} \\
    \midrule
    DeepLab v3+ & 235M & 639G \\
    DMRNet      & 238M & 639G \\
    Flex-MoE    & 672M & 716G  \\
    MoMKE       & 245M & 1297G \\
    SimMLM      & 247M & 955G  \\
    \cellcolor{gray!15}MARS        & \cellcolor{gray!15}244M & \cellcolor{gray!15}716G \\
    \bottomrule
  \end{tabular}
  }
  \vspace{0mm}
  \subcaption{Complexity analysis.}
  \label{tab:complexity}
\end{minipage}
\begin{minipage}[t]{0.29\columnwidth}
  \centering
  \vspace{-28.7mm}  
  \includegraphics[width=0.95\columnwidth]{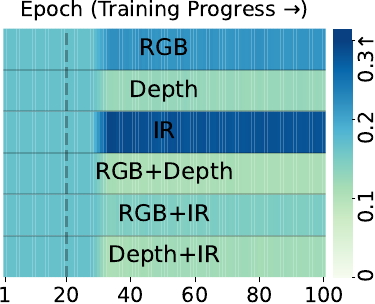}
  \vspace{0.8mm}
   \subcaption{Sampling analysis. }
   \label{fig:sampling}
\end{minipage}
\vspace{-3mm}
\caption{\small Comprehensive analyses of MARS. (a) Effect of the number of experts $N$ and top-$K$ on CASIA-SURF~\cite{CASIASURF}. (b) Parameter and FLOPs comparison on MCubeS~\cite{MCubeS}. (c) Sampling probabilities across modality combinations on CASIA-SURF~\cite{CASIASURF}.}
\vspace{-4mm}
\end{figure}

\vspace{-2pt}
\subsection{Comprehensive Analyses}

\noindent\textbf{Sensitivity Study.}
We analyze the sensitivity on the CASIA-SURF dataset~\cite{CASIASURF}.
In 
\cref{fig:sensitivity}, the top plot varies the number of experts with top-$K$ fixed to 5, 
and the bottom plot varies top-$K$ with the number of experts fixed to 16.
We observe that the performance degrades when either the number of experts or top-$K$ increases beyond the optimal setting.
This suggests that employing too many experts leads to redundancy among them, 
whereas activating more experts per sample could reduce their specialization.

\noindent{\bf Complexity Study.}
We evaluate model complexity on the MCubeS dataset~\cite{MCubeS} in terms of FLOPs and parameter count, as summarized in 
\cref{tab:complexity}.
We use segmentation to assess scalability, 
since its backbone (DeepLab~v3+~\cite{deeplab}) and resulting feature maps are heavier than those in classification tasks, 
providing a more realistic view of computational extensibility.
While Flex-MoE~\cite{FlexMoE} introduces substantial parameter overhead due to its modality bank, 
MARS remains compact by adding only a lightweight residual router.
Our FLOPs exhibit a moderate increase compared to DeepLab~v3+, 
which is expected from adopting the sparse MoE structure.
In contrast, other MoE-based frameworks, \ie, MoMKE~\cite{MoMKE} and SimMLM~\cite{SimMLM}, show considerably higher computational costs, \ie, FLOPs, due to cross-feeding encoders and logit-level aggregation, respectively.

   

\noindent\textbf{Sampling Study.}
\cref{fig:sampling} visualizes the sampling probabilities $q$ (in \cref{eq:sampling_prob}) across various modality combinations on the CASIA-SURF dataset~\cite{CASIASURF}.
During the first 20 epochs, sampling is performed uniformly for all combinations.
After activating the proposed discrepancy-aware sampling strategy (\cref{sec:sampling}), 
the IR-only configuration exhibits the highest probability ($\sim$0.43), 
indicating a large discrepancy $d$ between the feature and residual routers 
resulting in more emphasis during training. 
This aligns with the observation in \cref{sec:clf_result} that prior methods performed poorly under the IR-only setting,
while our model achieved significant improvement.
The next highest probabilities are observed for the RGB-only and RGB+IR combinations, 
confirming that the sampling mechanism prioritizes modality configurations requiring further refinement.

\noindent \textbf{Additional rationale for the residual} is provided in \cref{sec:justification}.



\section{Conclusion}
\vspace{-5pt}
In this paper, we present MARS,
a 
MoE 
framework that specializes experts by modeling missingness-induced representation deviations.
By comparing full- and partial-modality representations, the residual router exploits residual cues that reflect how missingness affects task reasoning, 
specializing experts for the resulting deviations.
The dual-router design and noise regularization together bridge the train–test routing gap, enabling benefits of residual-inspired routing even at inference.
Furthermore, the proposed sampling balances training across modality configurations.
Extensive incomplete multimodal experiments across domains and tasks demonstrate consistent superiority with minimal computational overhead.
Thus,
jointly leveraging observation and missingness
offers a principled path toward robust multimodal systems under test-time incompleteness.

\makeatletter
\let\addcontentsline\mars@origaddcontentsline
\makeatother



\newpage
\appendix

\renewcommand{\theHsection}{appendix.\Alph{section}}
\renewcommand{\theHsubsection}{appendix.\Alph{section}.\arabic{subsection}}
\renewcommand{\theHsubsubsection}{appendix.\Alph{section}.\arabic{subsection}.\arabic{subsubsection}}
\renewcommand{\theHfigure}{appendix.\arabic{figure}}
\renewcommand{\theHtable}{appendix.\arabic{table}}
\renewcommand{\theHequation}{appendix.\arabic{equation}}

\renewcommand\thefigure{A\arabic{figure}}
\renewcommand{\thetable}{A\arabic{table}}
\renewcommand{\theequation}{A\arabic{equation}}

\setcounter{figure}{0}
\setcounter{table}{0}
\setcounter{equation}{0}

\begin{center}
{\Large\bfseries Residual-Guided Expert Specialization \\for Incomplete Multimodal Learning\\[1.5mm]
\normalfont\textit{Supplementary Material}}
\end{center}

\makeatletter
\renewcommand{\tableofcontents}{
  \section*{\contentsname}
  \vspace{2pt}
  {
    \setlength{\parskip}{4pt}
    \@starttoc{toc}
  }
}
\makeatother

\vspace{3mm}
\centerline{\Large \textbf{Overview}}
\vspace{5mm}

\noindent$\blacktriangleright$ \textbf{Appendix Structure.}
To build intuition for the residuals, we provide further justification in \cref{sec:justification}.
The following sections (Sec.~\ref{sec:additional_experiment}--\ref{sec:details}) present additional experimental results and details that were omitted from the main paper due to space constraints.
We conclude the appendix with clarification of the technical novelty (\cref{sec:novelty}) and a discussion of limitations and future work (\cref{sec:limitation}).
\vspace{2mm}

\sloppy
\noindent$\blacktriangleright$ \textbf{Code Availability.}
We conduct experiments on CASIA-SURF~\cite{CASIASURF}, MCubeS~\cite{MCubeS}, CREMA-D~\cite{crema}, and UPMC~Food-101~\cite{food101} datasets.
The code and pretrained checkpoints are available at \url{https://github.com/seunghub/MARS}.

\vspace{5mm}
{\hypersetup{linkcolor=Black}
\setcounter{tocdepth}{2}
\begingroup
\let\clearpage\relax
\tableofcontents
\endgroup

}
\clearpage

\section{Additional Justification for Using Residual}
\label{sec:justification}

\begin{figure}[h]
  \centering
\vspace{-5mm}
    \includegraphics[width=0.99\columnwidth]{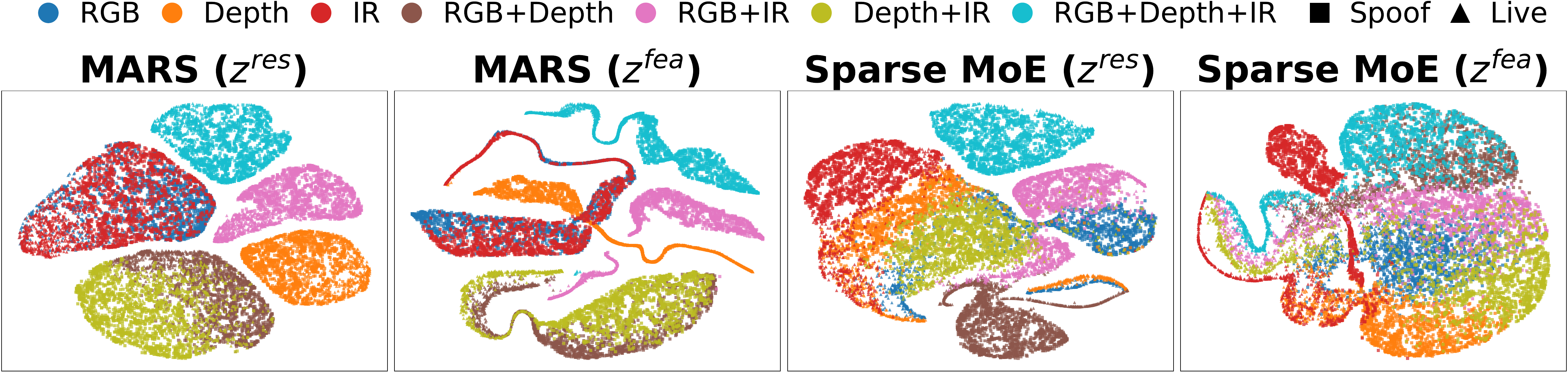}
    \vspace{-1.5mm}
   \caption{\footnotesize Embedding visualization from MARS and sparse MoE.}
   \vspace{-5mm}
   \label{fig:tsne}
\end{figure}

\subsection{Residual Analyses}
\label{sec:residual}
\cref{fig:tsne} visualizes feature $z^\text{fea}$ and residual embeddings $z^\text{res}$ across all configurations of 2,000 test samples using t-SNE.

\noindent\textbf{MARS:}
$z^{\text{res}}$ clusters by configuration with partial overlap, supporting our assumption. 
$z^{\text{fea}}$ follows this structure, where overlapping RGB/Depth/IR samples use same expert (e.g., E10 in Fig.~\textcolor{blue}{4}), indicating shared deviation patterns.
\\
\noindent\textbf{Sparse MoE:}
While $z^{\text{fea}}$ shows no clear clustering, 
constructing $z^{\text{res}}$ enables partial configuration-wise grouping.


\subsection{Interpreting Residuals in Deep Feature Space}

The linear difference (\ie, residual) between $z^\text{partial}$ and $z^\text{full}$ reflects the element-wise deviations of $z^\text{partial}$ from the most informative anchor $z^\text{full}$.
This allows the router to infer relative reliability without requiring explicit scaling or normalization, since such adjustments can be learned implicitly during training.
Although directly visualizing the residual $z^\text{res}$ is not straightforward, we've empirically demonstrated that this signal is effective in guiding expert specialization. 

Moreover, linear operations in learned representation spaces have been widely used in prior principled works.
For instance, Word2Vec~\cite{word2vec} demonstrates that vector offsets encode relational semantics, as illustrated by the classic example king - man + woman $\sim$ queen. 
Similarly, latent editing in GANs~\cite{gan1,gan2} shows that adding or subtracting latent directions produces interpretable semantic transformations such as pose, age, or expression. 
These indicate that linear operations within a shared manifold can induce meaningful changes in model behavior.

In our setting, both $z^\text{full}$ and $z^\text{partial}$ are generated by the same encoder and optimized for the same downstream objective. 
They therefore reside in a shared task-specific representation space. 
Their difference $z^\text{res}$ naturally reflects how the absence of modalities perturbs the evidence used by the model. 
From this perspective, the residual is not interpreted as a geometric distance but as a functional direction that isolates the task-relevant information removed by missing inputs. 
This interpretation aligns with a broad body of work showing that differences between deep representations can encode meaningful semantic or causal changes, even when the underlying feature space is highly entangled.


\begin{figure}[h]
  \centering
  \includegraphics[width=0.99\columnwidth]{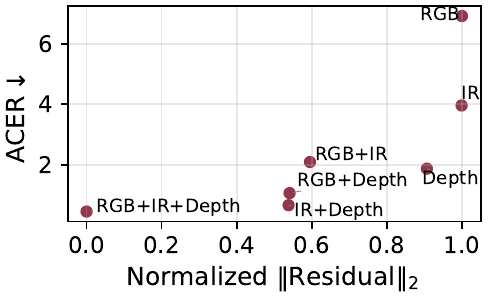}
  \vspace{-2mm}
   \caption{\small 
    Relationship between task performance and the normalized residual magnitude on CASIA-SURF~\cite{CASIASURF} at inference.
    Larger residuals, which reflect stronger representation deviations induced by missingness, generally correspond to higher error rates.
    }
\vspace{-2mm}
    
   \label{fig:residual_perforamcne}
\end{figure}

\subsection{Empirical Analysis of Representation Deviation}

As an additional post-hoc examination, we investigate the empirical relation between the residual and prediction accuracy. 
While the magnitude of the residual alone does not fully characterize the effect of modality missingness, since directional and feature-wise factors also contribute,
it provides a simple scalar summary that allows a coarse inspection of representation deviation.
As shown in Fig.~\ref{fig:residual_perforamcne}, larger residual magnitudes generally correspond to lower model accuracy across modality configurations.
This trend suggests that residuals capture representation shifts induced by modality missingness, and that their magnitude offers a coarse yet informative indicator of such deviations.
\vspace{-4mm}



\begin{table}[h]
\centering
\caption{\small Ablation study of privileged router inputs on multimodal classification (ACER ↓) using CASIA-SURF~\cite{CASIASURF}.
Alternative ways of combining complete and incomplete features consistently underperform compared to the residual formulation.
}
\renewcommand{\arraystretch}{0.96}
\resizebox{0.8\columnwidth}{!}{%
\begin{tabular}{c c c | c c c c c c c}
\toprule
\multicolumn{3}{c|}{Modality} & \multicolumn{7}{c}{Methods} \\
\midrule

RGB & Depth & IR & Sum & Product & Concat & Ratio & Cos Sim & Attention & \cellcolor{gray!15}MARS \\
\midrule
\CIRCLE & \Circle & \Circle & 10.31 & 10.62 & 8.64 & 25.74 & 10.81 & 11.86 & \cellcolor{gray!15}\textbf{6.92} \\
\Circle & \CIRCLE & \Circle & 2.67 & 3.82 & 1.72 & 5.22 & 4.02 & 2.34 & \cellcolor{gray!15}\textbf{1.87} \\
\Circle & \Circle & \CIRCLE & 7.90 & 8.50 & 9.40 & 21.92 & 9.29 & 8.59 & \cellcolor{gray!15}\textbf{3.96} \\
\midrule
\CIRCLE & \CIRCLE & \Circle & 1.98 & 2.38 & 1.08 & 4.01 & 1.98 & 2.01 & \cellcolor{gray!15}\textbf{1.06} \\
\CIRCLE & \Circle & \CIRCLE & 5.26 & 4.98 & 3.66 & 19.91 & 4.69 & 6.26 & \cellcolor{gray!15}\textbf{2.09} \\
\Circle & \CIRCLE & \CIRCLE & 1.47 & 2.79 & 1.17 & 3.74 & 2.58 & 1.18 & \cellcolor{gray!15}\textbf{0.66} \\
\midrule
\CIRCLE & \CIRCLE & \CIRCLE & 1.27 & 2.25 & 1.13 & 3.54 & 1.22 & 1.55 & \cellcolor{gray!15}\textbf{0.45} \\
\midrule
\multicolumn{3}{c|}{Average} & 4.41 & 5.05 & 3.83 & 12.01 & 4.94 & 4.83 & \cellcolor{gray!15}\textbf{2.43} \\
\bottomrule
\end{tabular}

}
\label{tab:classification}
\end{table}

\subsection{Comparative Analysis of Representation Deviation Modeling}
\vspace{-1mm}
A natural question is whether subtraction is the most suitable way to express the representation deviation introduced by modality missingness. 
To examine this, we compare it with several alternative formulations, including element-wise summation, product, ratio, cosine similarity, concatenation, and attention-based fusion. 
These alternatives capture different relationships between complete and incomplete representations, while concatenation additionally provides the router with both embeddings without imposing a predefined structure.

As shown in Table~\ref{tab:classification}, none of these alternatives outperforms the residual formulation. 
In the incomplete-modality setting, such operations often introduce additional interactions between the two representations, which can make the resulting signal more sensitive to the uncertainty present in incomplete embeddings. 
As a result, the router may receive less stable cues for expert selection.

Subtraction, in contrast, provides a direct characterization of the representation deviation caused by modality missingness. 
Its simplicity isolates the change between complete and incomplete embeddings without introducing additional interactions, yielding a clearer signal for routing decisions. 
This likely explains why the residual formulation consistently achieves the best performance.

\section{Additional Experiment}\label{sec:additional_experiment}
\begin{figure}[h]
  \centering
\vspace{-4mm}
  \includegraphics[width=0.90\columnwidth]{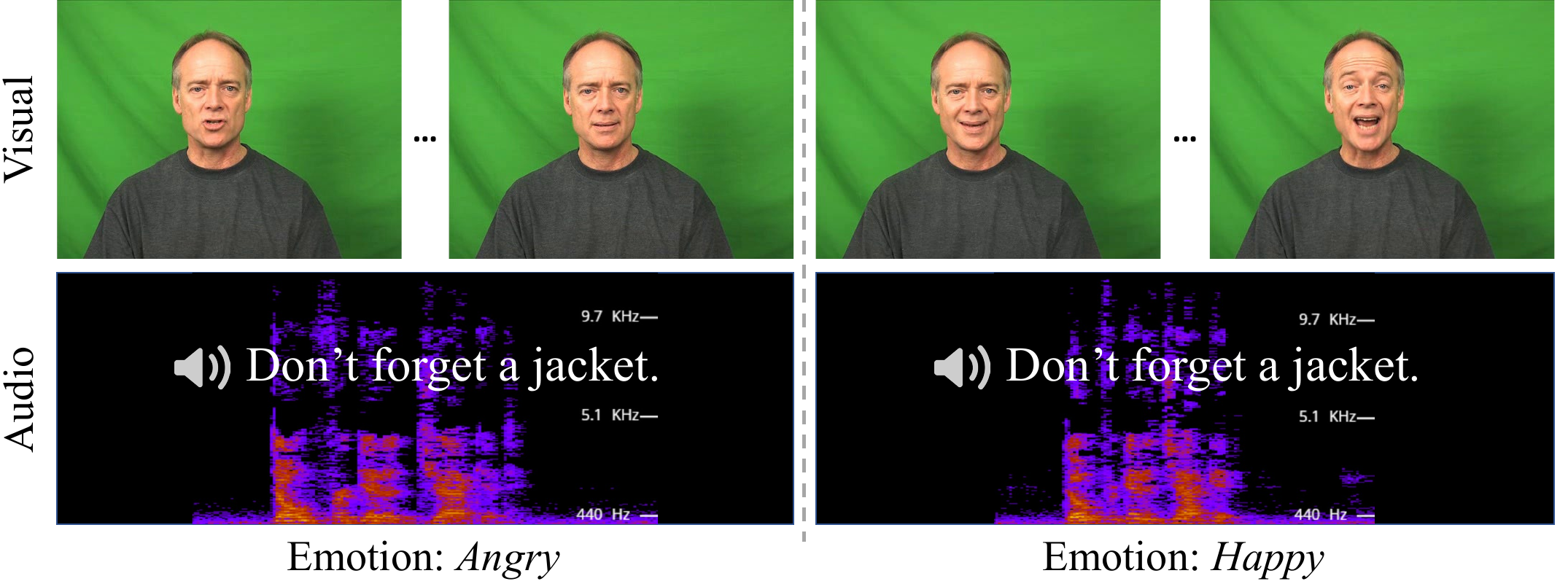}
   \caption{Visualization of a sample from the CREMA-D~\cite{crema} dataset.
    The same actor speaks the same sentence while expressing different emotions.
    Audio and visual modalities provide complementary cues, such as vocal tone and facial expressions.}
    \vspace{-4mm}
    
   \label{fig:cremad}
\end{figure}
\begin{figure}[h!]
  \centering
  \includegraphics[width=0.90\columnwidth]{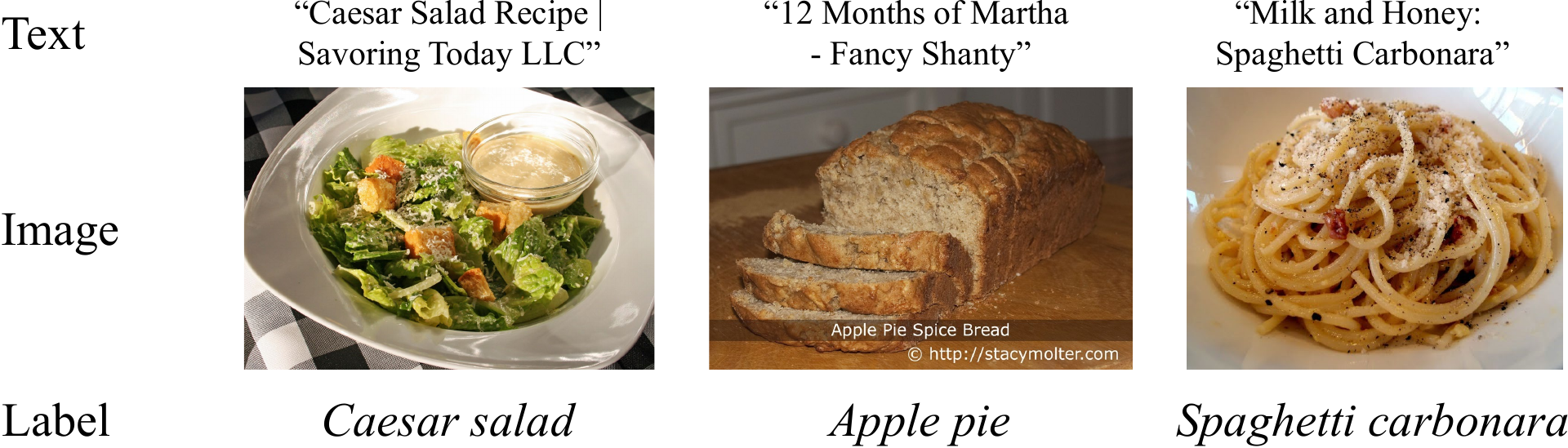}
    \caption{Visualization of samples from the UPMC~Food-101~\cite{food101} dataset.
    The text modality often provides strong cues by explicitly mentioning the food category (first and third samples),
    but in some cases it may be unrelated to the label (second sample).}
\label{fig:food101}
\end{figure}

\subsection{CREMA-D~\cite{crema} and UPMC~Food-101~\cite{food101} Dataset}

As visualized in \cref{fig:cremad}, CREMA-D~\cite{crema} is a multimodal emotion recognition dataset consisting of audio and visual recordings of acted emotional expressions.
It provides six categories (\emph{happy}, \emph{sad}, \emph{anger}, \emph{fear}, \emph{disgust}, and \emph{neutral}).
Following the benchmark~\cite{crema}, we use 6,698 samples for training and 744 samples for testing.

UPMC~Food-101~\cite{food101} is a multimodal food classification dataset composed of paired images and textual descriptions collected in uncontrolled environments.
It contains 101 food categories with 67,988 training samples and 22,716 test samples.
In \cref{fig:food101}, the textual descriptions often provide strong cues by directly mentioning the corresponding food category.
However, in some cases they contain only task-irrelevant information, as shown in the second sample of \cref{fig:food101}.

For both datasets, performance is evaluated using classification accuracy.

\subsection{Experimental Setups on CREMA-D~\cite{crema}}
We extend the reported performance of DMRNet~\cite{DMRNet} and apply the same task setting.
OGM-MD is a variant of OGM~\cite{OGM}, proposed as a baseline in DMRNet~\cite{DMRNet}, augmented with a binary indicator $\delta$ after each modality encoder.
ShaSpec~\cite{ShaSpec}, MMANet~\cite{ShaSpec}, and DMRNet~\cite{DMRNet} are included as recent baselines.

For consistency, we follow the model structure and data configuration of OGM~\cite{OGM}.
All remaining data preprocessing and training procedures adhere to the settings described in DMRNet~\cite{DMRNet}.
As in our other experiments, 
we activate the top-5 experts out of 16.
The residual router generates clean logits and noise through two separate linear projection layers, 
while the feature router is implemented as a 2-layer MLP.
The weights $\lambda_\text{task}$, $\lambda_\text{LB}$, $\lambda_\text{distill}$, and $\lambda_\text{noise}$ are set to 1, 1, 5, and 1, respectively.
Our sampling strategy starts from epoch 20.

\subsection{Experimental Setups on UPMC~Food-101~\cite{food101}}
For UPMC~Food-101, we reproduce the experimental setting of SimMLM~\cite{SimMLM}. 
Since the official implementation is not publicly available, we follow the model description provided in their paper. 
The image is encoded using Inception~V3~\cite{inception}, while text is processed with BERT~\cite{bert}. 
Image and text representations are fused using the concatenation scheme proposed in the original dataset work~\cite{food101}.

In this setup, we employ a smaller MoE configuration with eight experts and top-3 routing. 
Each expert is implemented as a two-layer MLP operating on the concatenated representation. 
The residual router generates routing logits and noise using two linear projection layers, and the feature router is implemented as a two-layer MLP. 
The loss weights $\lambda_\text{task}$, $\lambda_\text{LB}$, $\lambda_\text{distill}$, and $\lambda_\text{noise}$ are set to 1, 0.05, 0.5, and 0.01, respectively. 
The model is trained for 100 epochs, and the proposed sampling mechanism starts from epoch 3.
We use the AdamW~\cite{adamw} optimizer with a learning rate of $1$$\times$$10^{-3}$ and a weight decay of $1$$\times$$10^{-4}$.
A smaller learning rate is applied to the Inception~V3~\cite{inception} and BERT~\cite{bert} encoders.
The learning rate is reduced by a factor of 0.1 when the loss plateaus.




\subsection{Result Analysis (Tab.~\textcolor{blue}{3} and Tab.~\textcolor{blue}{4} in the main paper)}
On the CREMA-D~\cite{crema} dataset, DMRNet~\cite{DMRNet} achieves the strongest performance among the baselines, consistent with observations from our main experiments.
MARS further improves the average accuracy by a clear margin of $\sim$4.17\%p over DMRNet.
These consistent gains in the audio–visual setting demonstrate the generality and effectiveness of MARS across different multimodal domains.


On the UPMC~Food-101~\cite{food101} dataset, MARS attains the best average accuracy of 91.59\%, with its advantage concentrated in the missing-modality configurations rather than the full-modality setting.
This improvement is particularly meaningful for this dataset, where the textual modality exhibits highly variable informativeness.
Descriptions may either explicitly reveal the food category or contain largely task-irrelevant or even misleading content.
The results suggest that residual-guided routing enables the model to account for task-specific representational deviations rather than relying on predefined expert roles (i.e., modality configurations).
As a result, the model maintains stable performance even when certain modalities provide noisy or unreliable cues, remaining above 90\% regardless of which modality is missing in this benchmark.

\clearpage
\begin{figure*}[!t]
  \centering
  \includegraphics[width=0.99\linewidth]{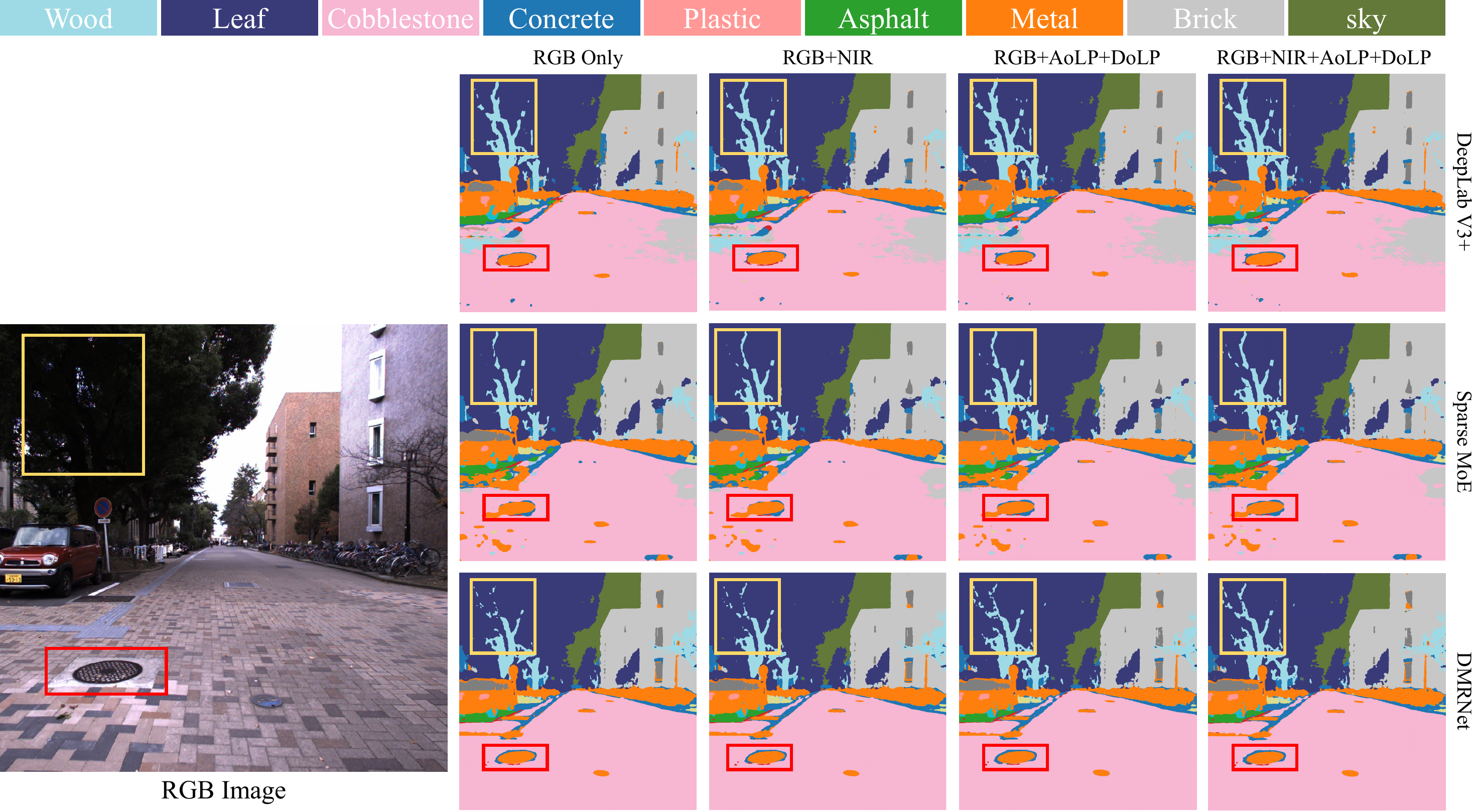}
    \vskip 0.5mm
  \includegraphics[width=0.99\linewidth]{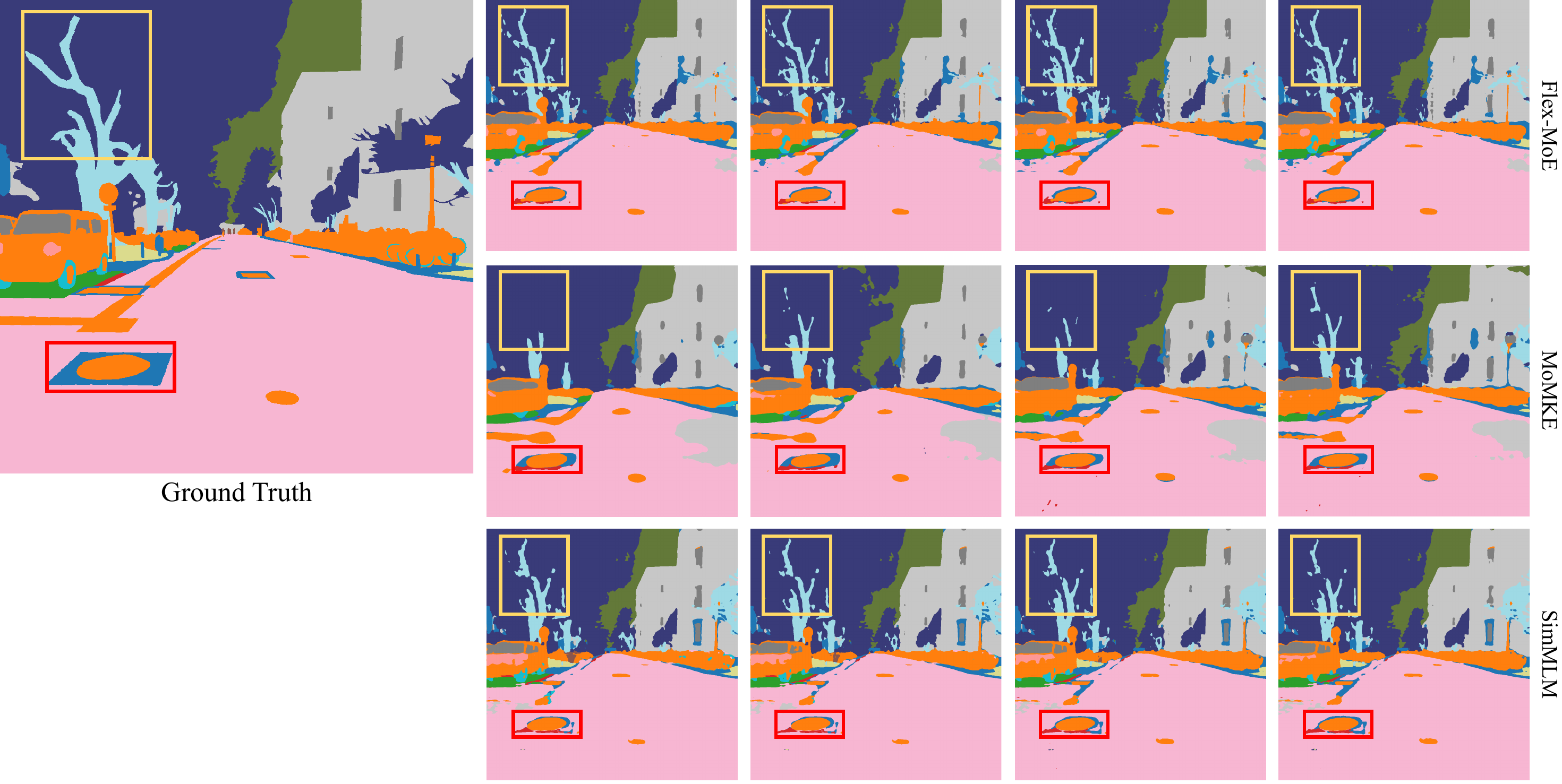}
    \vskip -0.4mm
  \includegraphics[width=0.99\linewidth]{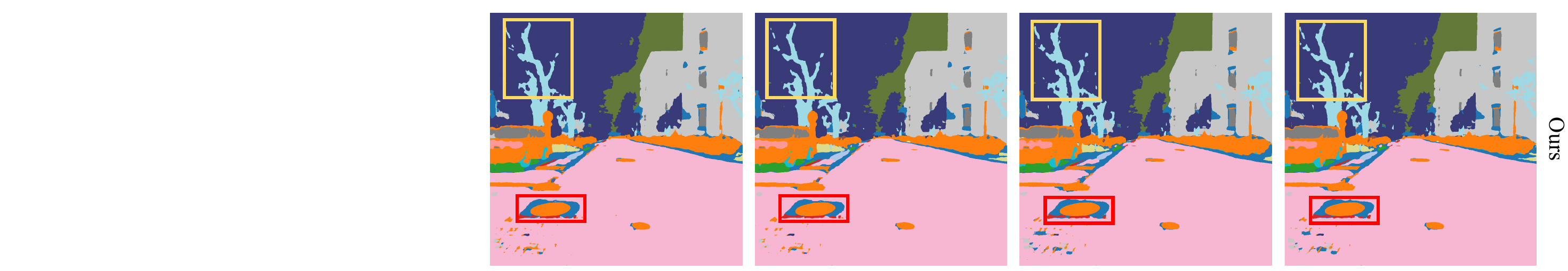}
   \caption{Qualitative results on multimodal segmentation using MCubeS~\cite{MCubeS}. Compared to the baselines, MARS better separates concrete from cobblestone (red box) and wood from leaves (yellow box), even when only the RGB modality is available.}
   \label{fig:qualitative_result}
\end{figure*}

\vspace{-4mm}
\section{Additional Qualitative Result}
\label{sec:qualitative}
\vspace{-1mm}
In \cref{fig:qualitative_result}, we provide additional qualitative examples from the MCubeS dataset~\cite{MCubeS}, comparing our method with all baselines.
Since Near-Infrared (NIR) and polarization cues (AoLP and DoLP) offer complementary information beyond RGB, leveraging more modalities generally leads to improved segmentation performance (see RGB Only vs.\ RGB+NIR+AoLP+DoLP).
When only RGB is available, however, differentiating concrete from cobblestone (red box) and woods from leaves (yellow box) remains difficult.
Although certain baselines occasionally perform well on one of these regions, none of them consistently handle both.
In contrast, our method produces reliable predictions across both challenging areas, demonstrating stronger robustness under incomplete modalities.

\begin{figure*}[!h]
  \centering
  \includegraphics[width=0.99\linewidth]{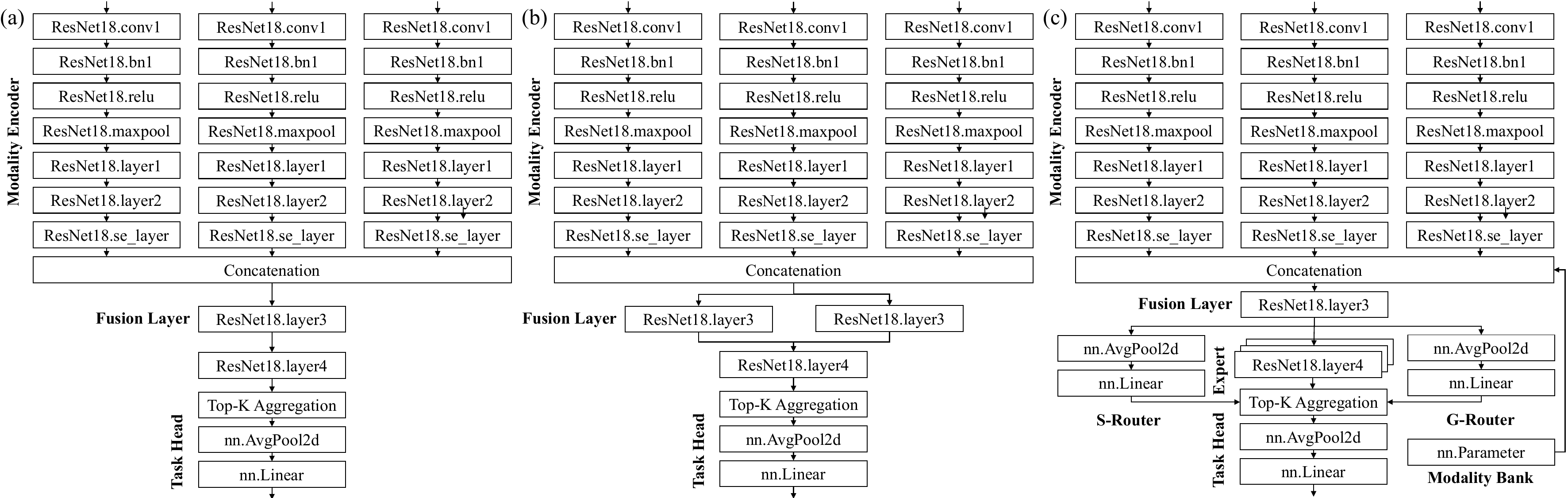}
    \vskip 3mm
  \includegraphics[width=0.99\linewidth]{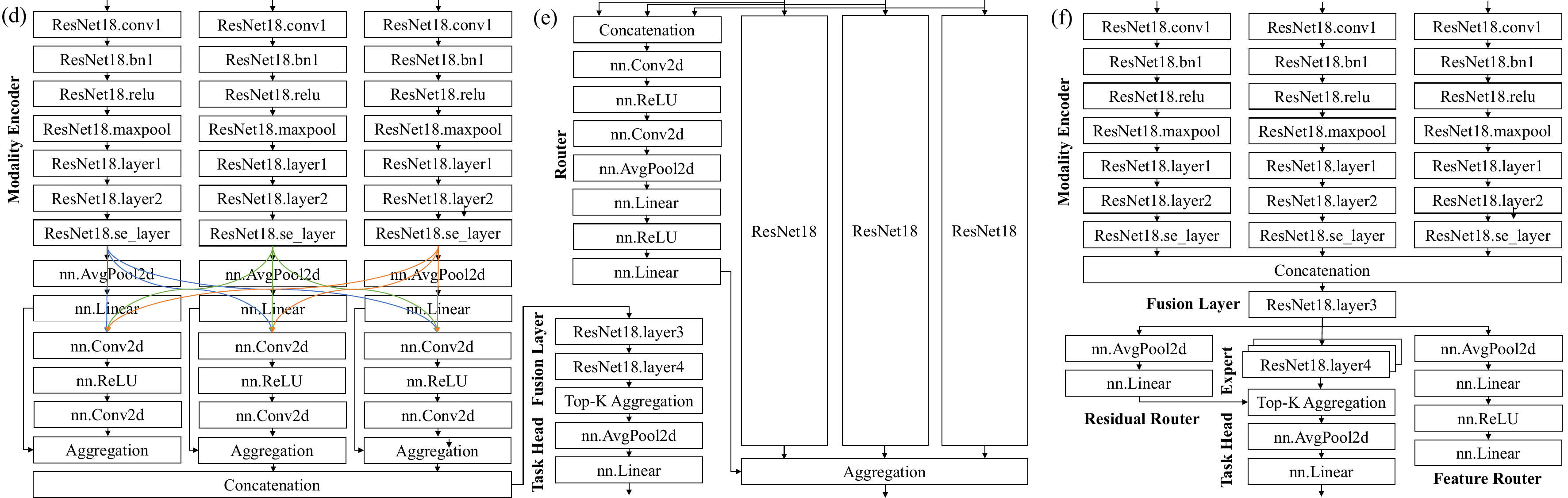}
   \caption{Architectural overview of the baseline models and the proposed method on CASIA-SURF~\cite{CASIASURF}: (a) ResNet-18~\cite{resnet}, (b) DMRNet~\cite{DMRNet}, (c) Flex-MoE~\cite{FlexMoE}, (d) MoMKE~\cite{MoMKE}, (e) SimMLM~\cite{SimMLM}, and (f) MARS (ours).}
   \label{fig:model_architecture}
\end{figure*}
\vspace{-10mm}

\begin{table*}[!h]
\centering
\caption{
Comprehensive training configurations for all compared methods on CASIA-SURF~\cite{CASIASURF} (left) and MCubeS~\cite{MCubeS} (right).}
\renewcommand{\arraystretch}{1.15}
\resizebox{\textwidth}{!}{
\begin{tabular}{l | c c c c >{\columncolor{gray!15}}c | c c c c >{\columncolor{gray!15}}c}
\toprule
& \multicolumn{5}{c|}{\textbf{CASIA-SURF}} & \multicolumn{5}{c}{\textbf{MCubeS}} \\
\midrule
\textbf{Setting} 
& \textbf{DMRNet} & \textbf{Flex-MoE} & \textbf{MoMKE} & \textbf{SimMLM} & \textbf{MARS} 
& \textbf{DMRNet} & \textbf{Flex-MoE} & \textbf{MoMKE} & \textbf{SimMLM} & \textbf{MARS} \\
\midrule
Learning rate & $1$$\times$$10^{-3}$ & $1$$\times$$10^{-2}$ & $1$$\times$$10^{-5}$ & $5$$\times$$10^{-3}$ & $5$$\times$$10^{-4}$& $5$$\times$$10^{-2}$ & $5$$\times$$10^{-3}$ & $5$$\times$$10^{-3}$ & $5$$\times$$10^{-2}$ & $5$$\times$$10^{-3}$ \\
\midrule
Weight decay & \multicolumn{5}{c|}{$5$$\times$$10^{-4}$} & \multicolumn{5}{c}{$5$$\times$$10^{-4}$} \\
Epochs & \multicolumn{5}{c|}{100} & \multicolumn{5}{c}{500} \\
Batch size & \multicolumn{5}{c|}{64} & \multicolumn{5}{c}{8} \\
Optimizer & \multicolumn{5}{c|}{SGD} & \multicolumn{5}{c}{SGD} \\
Backbone & \multicolumn{5}{c|}{ResNet-18} & \multicolumn{5}{c}{DeepLab~V3+} \\
Augmentation & \multicolumn{5}{c|}{Horizontal Flip, Random Crop, Random Rotation} &  \multicolumn{5}{c}{Horizontal Flip, Random Crop, Gaussian Blur} \\

\bottomrule
\end{tabular}
}
\label{tab:impl_details}
\end{table*}

\section{Additional Experimental Details}
\vspace{-2mm}
\label{sec:details}
As illustrated in \cref{fig:model_architecture}, our model and baselines are implemented on top of the same backbone architecture.
Model-specific variations follow their official codebases or papers.
We conduct an extensive hyperparameter search for all baselines to obtain their best performance, and the selected configurations are summarized in \cref{tab:impl_details}.
Detailed setups for each baseline are provided below.

\vspace{-2mm}

\subsection{Experimental Setups in CASIA-SURF~\cite{CASIASURF}}\label{sec:casia-detail}
\noindent\textbf{DMRNet~\cite{DMRNet}.}
We report the presented results in the paper.

\sloppy
\noindent\textbf{Flex-MoE~\cite{FlexMoE}.}
Flex-MoE is originally designed for training scenarios with incomplete modalities.
To align it with our setting and achieve its best performance, we provide full-modality inputs for training the $\mathcal{G}$-router, and masked-modality inputs for training the $\mathcal{S}$-router.
This configuration offers a more privileged setting than their original formulation but matches our IML setup.
Both routers use a single projection layer, and the missing-modality bank operates on $z^{(m)}i$.
We found that selecting Top-5 experts out of 16 yielded the best performance within the search ranges of $K\in\{3,4,5,6\}$ and $N\in\{8,16,32\}$.
The optimal balancing-loss weight was $\mathcal{L}_{LB}$$=$$0.1$ (searched over $\{0.5,0.1,0.05,0.01\}$), and the optimal top-1 guiding loss weight was $\mathcal{L}_\text{ce}$$=$$0.1$ (searched over $\{1,0.5,0.1,0.05,0.01,0.005,0.001\}$).

\noindent\textbf{MoMKE~\cite{MoMKE}.}
We first train an individual ResNet-18~\cite{resnet} for each modality for 50 epochs with a learning rate of $1$$\times$$10^{-3}$.
We then train the full model with the experts and a projection router for 100 epochs with a learning rate of $5$$\times$$10^{-4}$.
We search learning rates over $[10^{-4},10^{-2}]$ and training epochs over $[50,200]$.

\noindent\textbf{SimMLM~\cite{SimMLM}.}
Following MoMKE, we train modality-specific ResNet-18 models for 50 epochs with a learning rate of $1$$\times$$10^{-3}$.
We then attach a lightweight router network that produces aggregation weights for modality-specific outputs.
The second stage is trained for 100 epochs with a learning rate of $5$$\times$$10^{-3}$,
using the same hyperparameter search space as MoMKE.
For the MoFe regularization, the best performance is achieved with weight $0.1$ (searched over ${1,0.5,0.1,0.05,0.01}$).

\vspace{-2mm}
\subsection{Experimental Setups in MCubeS~\cite{MCubeS}}

Unless otherwise specified, we reuse the configurations and hyperparameter search ranges described in Section~\ref{sec:casia-detail}.

\noindent\textbf{DMRNet~\cite{DMRNet}.}
Following their formulation, we replace $z_i$ with $\mu_i + \epsilon\sigma_i$,
where $\mu_i$ and $\sigma_i$ are learned in the same manner as $z_i$,
and $\epsilon\sim\mathcal{N}(0,I)$.
We set the distribution-regularization weight to $\mathcal{L}_\text{DR}$$=$$1$$\times$$10^{-4}$ (searched over $\{10^{-2},10^{-3},10^{-4},10^{-5}\}$) and the hard-combination regularization weight to $\mathcal{L}_\text{HCR}$$=$$0.1$ (searched over $\{1,0.5,0.1,0.05,0.01,0.005,0.001\}$).

\noindent\textbf{Flex-MoE~\cite{FlexMoE}.}
We adopt the same strategy as in CASIA-SURF~\cite{CASIASURF} (Section~\ref{sec:casia-detail}),
but the optimal balancing-loss weight was $\mathcal{L}\text{LB}=0.01$
and the top-1 guiding loss weight was $\mathcal{L}\text{ce}=0.005$ within the same search range.

\noindent\textbf{MoMKE~\cite{MoMKE}.}
We train the first and second stages for 200 and 500 epochs with learning rates of $1\times10^{-4}$ and $5\times10^{-3}$, respectively.
We use a 2-layer CNN cross-modal encoder and a 2-layer MLP router with average pooling.

\noindent\textbf{SimMLM~\cite{SimMLM}.}
We use the same first-stage setup as in MoMKE~\cite{MoMKE}.
In the second stage, we train for 500 epochs with a learning rate of $5$$\times$$ 10^{-2}$.
The router consists of a 2-layer CNN followed by pooling and a linear layer,
and $\mathcal{L}_\text{MoFe}$$=$$0.1$.


\vspace{-2mm}
\section{Technical Novelty Clarification}
\label{sec:novelty}
\vspace{-1mm}
To the best of our knowledge, the technical novelty of MARS lies in three aspects.
\vspace{-4mm}

\begin{itemize}

\item \textbf{Modeling Representation Deviations.}
Most prior IML methods focus on making better use of the simulated inputs, for example through modality imputation, distillation, or robust feature fusion.
MARS instead focuses on explicitly modeling how the task representation changes when modalities are missing.
Our key idea is that incomplete modalities inherently produce imperfect task representations compared to full-modality inputs, even with sophisticated modeling.
When modalities contain complementary information, their absence inevitably introduces representational deviations.
By explicitly capturing this deviation through the residual between complete and incomplete embeddings, our framework shifts the perspective of IML from missing-data compensation to representation-deviation modeling.
\vspace{2mm}

\item \textbf{Privileged Signals for Routing Specialization.}
MARS is related to the Learning Using Privileged Information (LUPI) paradigm~\cite{LUPI1,LUPI2,LUPI3} in that additional information is available during training but not at inference.
In MARS, privileged signals (i.e., $z^\text{full}$) are used to compute residual representations that guide \emph{expert specialization}.
This design decouples expert specialization from deployment routing. 
It introduces routing mechanisms with different inputs for specialization and deployment.
Such a routing-oriented use of privileged signals has not been explored in prior studies.
\vspace{2mm}

\item \textbf{Task-Oriented Routing Noise.}
Conventional noisy routing in mixture-of-experts models mainly serves to improve expert diversity.
In contrast, MARS leverages routing noise to address the discrepancy between the privileged residual router and the deployable feature router.
Expert activation is therefore influenced through two complementary channels as deterministic routing logits and discrepancy-aware stochastic modulation.
This perspective expands the role of routing noise from a generic regularizer to a controllable signal that stabilizes routing under train--test modality gaps.

\end{itemize}

\vspace{-5mm}
\section{Limitation and Future Work}
\vspace{-1mm}
\label{sec:limitation}
MARS is designed for the standard IML setting where {\em all modalities are available during training} and missingness occurs only at inference.
Under this assumption, the residual signals are reliably constructed from complete inputs, which enables effective expert specialization.
However, in {\em scenarios where the training data themselves contain incomplete modalities}, the residual supervision may not be directly obtainable.
Adopting our approach to such settings would require redefining how to estimate representation deviations without consistent full-modality references.
We leave this as an interesting direction for future work.


\section*{Acknowledgements}
This research was supported by 
RS-2025-02216257 (65\%),
RS-2022-II220290 (30\%), and 
RS-2019-II1091906 (AI Graduate Program at POSTECH, 5\%).

%
%
\bibliographystyle{splncs04}
\bibliography{main}

@inproceedings{MMANet,
  title={Mmanet: Margin-aware distillation and modality-aware regularization for incomplete multimodal learning},
  author={Wei, Shicai and Luo, Chunbo and Luo, Yang},
  booktitle={CVPR},
  pages={20039--20049},
  year={2023}
}

@inproceedings{HeMIS,
  title={Hemis: Hetero-modal image segmentation},
  author={Havaei, Mohammad and Guizard, Nicolas and Chapados, Nicolas and Bengio, Yoshua},
  booktitle={MICCAI},
  pages={469--477},
  year={2016},
  organization={Springer}
}

@inproceedings{LCR,
  title={Brain tumor segmentation with missing modalities via latent multi-source correlation representation},
  author={Zhou, Tongxue and Canu, St{\'e}phane and Vera, Pierre and others},
  booktitle={MICCAI},
  pages={533--541},
  year={2020},
  organization={Springer}
}

@inproceedings{RFNet,
  title={RFNet: Region-aware fusion network for incomplete multi-modal brain tumor segmentation},
  author={Ding, Yuhang and Yu, Xin and Yang, Yi},
  booktitle={ICCV},
  pages={3975--3984},
  year={2021}
}

@inproceedings{mmformer,
  title={mmformer: Multimodal medical transformer for incomplete multimodal learning of brain tumor segmentation},
  author={Zhang, Yao and He, Nanjun and Yang, Jiawei and others},
  booktitle={MICCAI},
  pages={107--117},
  year={2022},
  organization={Springer}
}

@inproceedings{ShaSpec,
  title={Multi-modal learning with missing modality via shared-specific feature modelling},
  author={Wang, Hu and Chen, Yuanhong and Ma, Congbo and others},
  booktitle={CVPR},
  pages={15878--15887},
  year={2023}
}

@inproceedings{DMRNet,
  title={Robust multimodal learning via representation decoupling},
  author={Wei, Shicai and Luo, Yang and Wang, Yuji and others},
  booktitle={ECCV},
  pages={38--54},
  year={2024},
  organization={Springer}
}

@inproceedings{MoMKE,
  title={Leveraging knowledge of modality experts for incomplete multimodal learning},
  author={Xu, Wenxin and Jiang, Hexin and Liang, Xuefeng},
  booktitle={ACM MM},
  pages={438--446},
  year={2024}
}

@inproceedings{SimMLM,
  title={SimMLM: A Simple Framework for Multi-modal Learning with Missing Modality},
  author={Li, Sijie and Chen, Chen and Han, Jungong},
  booktitle={ICCV},
  year={2025}
}

@article{sparseMoE,
  title={Outrageously large neural networks: The sparsely-gated mixture-of-experts layer},
  author={Shazeer, Noam and Mirhoseini, Azalia and Maziarz, Krzysztof and others},
  journal={ICLR},
  year={2017}
}

@article{FlexMoE,
  title={Flex-moe: Modeling arbitrary modality combination via the flexible mixture-of-experts},
  author={Yun, Sukwon and Choi, Inyoung and Peng, Jie and others},
  journal={NeurIPS},
  volume={37},
  pages={98782--98805},
  year={2024}
}

@inproceedings{CASIASURF,
  title={A dataset and benchmark for large-scale multi-modal face anti-spoofing},
  author={Zhang, Shifeng and Wang, Xiaobo and Liu, Ajian and others},
  booktitle={CVPR},
  pages={919--928},
  year={2019}
}

@inproceedings{MCubeS,
  title={Multimodal material segmentation},
  author={Liang, Yupeng and Wakaki, Ryosuke and Nobuhara, Shohei and others},
  booktitle={CVPR},
  pages={19800--19808},
  year={2022}
}

@article{hallucination,
  title={AI hallucination: towards a comprehensive classification of distorted information in artificial intelligence-generated content},
  author={Sun, Yujie and Sheng, Dongfang and Zhou, Zihan and others},
  journal={Humanities and Social Sciences Communications},
  volume={11},
  number={1},
  pages={1--14},
  year={2024},
  publisher={Palgrave}
}

@inproceedings{hallucination2,
  title={Hallucination index: An image quality metric for generative reconstruction models},
  author={Tivnan, Matthew and Yoon, Siyeop and Chen, Zhennong and others},
  booktitle={MICCAI},
  pages={449--458},
  year={2024},
  organization={Springer}
}

@article{multimodal_segmentation,
  title={Deep multimodal fusion for semantic image segmentation: A survey},
  author={Zhang, Yifei and Sidib{\'e}, D{\'e}sir{\'e} and Morel, Olivier and others},
  journal={Image and Vision Computing},
  volume={105},
  pages={104042},
  year={2021},
  publisher={Elsevier}
}

@article{multimodal_classification,
  title={Multimodal classification: Current landscape, taxonomy and future directions},
  author={Sleeman IV, William C and Kapoor, Rishabh and Ghosh, Preetam},
  journal={ACM Computing Surveys},
  volume={55},
  number={7},
  pages={1--31},
  year={2022},
  publisher={ACM New York, NY}
}

@article{remotesensing1,
  title={Multimodal classification of remote sensing images: A review and future directions},
  author={G{\'o}mez-Chova, Luis and Tuia, Devis and Moser, Gabriele and others},
  journal={Proceedings of the IEEE},
  volume={103},
  number={9},
  pages={1560--1584},
  year={2015},
  publisher={IEEE}
}

@article{remotesensing2,
  title={A multilevel multimodal fusion transformer for remote sensing semantic segmentation},
  author={Ma, Xianping and Zhang, Xiaokang and Pun, Man-On and others},
  journal={IEEE Transactions on Geoscience and Remote Sensing},
  volume={62},
  pages={1--15},
  year={2024},
  publisher={IEEE}
}

@article{medical3,
  title={A review of deep learning-based information fusion techniques for multimodal medical image classification},
  author={Li, Yihao and Daho, Mostafa El Habib and Conze, Pierre-Henri and Zeghlache, Rachid and Le Boit{\'e}, Hugo and Tadayoni, Ramin and Cochener, B{\'e}atrice and Lamard, Mathieu and Quellec, Gwenol{\'e}},
  journal={Computers in Biology and Medicine},
  volume={177},
  pages={108635},
  year={2024},
  publisher={Elsevier}
}

@article{survey,
  title={Deep multimodal learning with missing modality: A survey},
  author={Wu, Renjie and Wang, Hu and Chen, Hsiang-Ting and others},
  journal={ACM Computing Surveys},
  year={2024}
}

@inproceedings{deeplab,
  title={Encoder-decoder with atrous separable convolution for semantic image segmentation},
  author={Chen, Liang-Chieh and Zhu, Yukun and Papandreou, George and others},
  booktitle={ECCV},
  pages={801--818},
  year={2018}
}

@inproceedings{resnet,
  title={Deep residual learning for image recognition},
  author={He, Kaiming and Zhang, Xiangyu and Ren, Shaoqing and Sun, Jian},
  booktitle={CVPR},
  pages={770--778},
  year={2016}
}

@inproceedings{gradcam,
  title={Grad-cam: Visual explanations from deep networks via gradient-based localization},
  author={Selvaraju, Ramprasaath R and Cogswell, Michael and Das, Abhishek and Vedantam, Ramakrishna and others},
  booktitle={ICCV},
  pages={618--626},
  year={2017}
}

@inproceedings{imputation1,
  title={Deep adversarial learning for multi-modality missing data completion},
  author={Cai, Lei and Wang, Zhengyang and Gao, Hongyang and others},
  booktitle={ACM SIGKDD},
  pages={1158--1166},
  year={2018}
}

@article{imputation3,
  title={Rethinking the diffusion models for missing data imputation: A gradient flow perspective},
  author={Chen, Zhichao and Li, Haoxuan and Wang, Fangyikang and others},
  journal={NeurIPS},
  volume={37},
  pages={112050--112103},
  year={2024}
}

@article{sgd,
  title={An overview of gradient descent optimization algorithms},
  author={Ruder, Sebastian},
  journal={arXiv preprint arXiv:1609.04747},
  year={2016}
}

@inproceedings{multimodal_classification2,
  title={Multi-level confidence learning for trustworthy multimodal classification},
  author={Zheng, Xiao and Tang, Chang and Wan, Zhiguo and others},
  booktitle={AAAI},
  volume={37},
  number={9},
  pages={11381--11389},
  year={2023}
}

@article{crema,
  title={Crema-d: Crowd-sourced emotional multimodal actors dataset},
  author={Cao, Houwei and Cooper, David G and Keutmann, Michael K and others},
  journal={IEEE transactions on affective computing},
  volume={5},
  number={4},
  pages={377--390},
  year={2014},
  publisher={IEEE}
}

@inproceedings{food101,
  title={Image and text fusion for upmc food-101 using bert and cnns},
  author={Gallo, Ignazio and Ria, Gianmarco and Landro, Nicola and La Grassa, Riccardo},
  booktitle={2020 35th International conference on image and vision computing New Zealand (IVCNZ)},
  pages={1--6},
  year={2020},
  organization={IEEE}
}

@inproceedings{train_missing,
  title={Are multimodal transformers robust to missing modality?},
  author={Ma, Mengmeng and Ren, Jian and Zhao, Long and others},
  booktitle={CVPR},
  pages={18177--18186},
  year={2022}
}

@inproceedings{medical4,
  title={Learning covariance-based multi-scale representation of neuroimaging measures for alzheimer classification},
  author={Baek, Seunghun and Choi, Injun and others},
  booktitle={ISBI},
  pages={1--5},
  year={2023},
  organization={IEEE}
}

@inproceedings{medical5,
  title={Ocl: Ordinal contrastive learning for imputating features with progressive labels},
  author={Baek, Seunghun and Sim, Jaeyoon and Wu, Guorong and Kim, Won Hwa},
  booktitle={MICCAI},
  pages={334--344},
  year={2024},
  organization={Springer}
}

@inproceedings{OGM,
  title={Balanced multimodal learning via on-the-fly gradient modulation},
  author={Peng, Xiaokang and Wei, Yake and Deng, Andong and others},
  booktitle={CVPR},
  pages={8238--8247},
  year={2022}
}

@article{word2vec,
  title={Efficient estimation of word representations in vector space},
  author={Mikolov, Tomas and Chen, Kai and Corrado, Greg and others},
  journal={arXiv preprint arXiv:1301.3781},
  year={2013}
}

@article{gan1,
  title={Interfacegan: Interpreting the disentangled face representation learned by gans},
  author={Shen, Yujun and Yang, Ceyuan and Tang, Xiaoou and others},
  journal={TPAMI},
  volume={44},
  number={4},
  pages={2004--2018},
  year={2020},
  publisher={IEEE}
}

@article{gan2,
  title={Ganspace: Discovering interpretable gan controls},
  author={H{\"a}rk{\"o}nen, Erik and Hertzmann, Aaron and Lehtinen, Jaakko and others},
  journal={NeurIPS},
  volume={33},
  pages={9841--9850},
  year={2020}
}

@inproceedings{inception,
  title={Rethinking the inception architecture for computer vision},
  author={Szegedy, Christian and Vanhoucke, Vincent and Ioffe, Sergey and others},
  booktitle={CVPR},
  pages={2818--2826},
  year={2016}
}

@inproceedings{bert,
  title={Bert: Pre-training of deep bidirectional transformers for language understanding},
  author={Devlin, Jacob and Chang, Ming-Wei and Lee, Kenton and others},
  booktitle={NAACL},
  pages={4171--4186},
  year={2019}
}

@article{adamw,
  title={Decoupled weight decay regularization},
  author={Loshchilov, Ilya and Hutter, Frank},
  journal={ICLR},
  year={2017}
}

@article{LUPI1,
  title={Learning using privileged information: similarity control and knowledge transfer},
  author={Vapnik, Vladimir and Izmailov, Rauf},
  journal={JMLR},
  volume={16},
  number={1},
  pages={2023--2049},
  year={2015},
}

@article{LUPI2,
  title={A new learning paradigm: Learning using privileged information},
  author={Vapnik, Vladimir and Vashist, Akshay},
  journal={Neural networks},
  volume={22},
  number={5-6},
  pages={544--557},
  year={2009},
  publisher={Elsevier}
}

@inproceedings{LUPI3,
  title={Learning to rank using privileged information},
  author={Sharmanska, Viktoriia and Quadrianto, Novi and others},
  booktitle={ICCV},
  pages={825--832},
  year={2013}
}

\end{document}